\pdfoutput=1
\documentclass[11pt]{article}

\usepackage[preprint]{acl}
\usepackage{times}
\usepackage{latexsym}
\usepackage{amsmath}
\usepackage{booktabs}
\usepackage[T1]{fontenc}

\usepackage{xspace}
\usepackage[utf8]{inputenc}
\usepackage{xcolor}
\usepackage[normalem]{ulem}
\PassOptionsToPackage{capitalize,nameinlink}{cleveref}
\usepackage{cleveref}

\usepackage{microtype}
\newcommand{\biasname}{search-by-answer bias\xspace}
\newcommand{\objchange}{object-contrast\xspace}
\newcommand{\relchange}{relation–object conflict\xspace}

\usepackage{inconsolata}

\usepackage{graphicx}
\usepackage{subcaption}
\usepackage{dblfloatfix}

\usepackage{url}
\usepackage{natbib}

\usepackage{algorithm}
\usepackage{algpseudocode}

\title{Search-by-Answer Bias in Retrieval Models}

\author{
Adiel Avigad \\
Bar-Ilan University \\
\texttt{avigada@biu.ac.il}
\And
Shauli Ravfogel \\
NYU Center for Data Science \\
\texttt{shauli.ravfogel@nyu.edu}
\And
Yoav Goldberg \\
Bar-Ilan University \\
\texttt{yogo@cs.biu.ac.il}
}
\usepackage{xcolor}
\begin{document}
\maketitle
\begin{abstract}
Dense retrievers are meant to select passages that match a query's
semantic specification. We show they suffer from \biasname{}: they favor
passages that mention answer objects the model knows (scores as likely), 
over passages that support valid alternative answers. We test this with controlled factual contrasts from LAMA/TREx. For each
retriever, we measure how strongly the model scores the canonical object
over an alternative object using a token-prediction head, and ask
whether this score predicts passage rankings. Across eight
retrievers---DPR, Spider variants, Contriever, E5, and Qwen3-Embedding---
larger token-prediction scores consistently predict higher retrieval
scores for passages containing the corresponding object. This holds when
passages differ only in the answer object, and even when the passage that
mentions the high-scoring object expresses a wrong relation. Dense
retrieval thus exhibits \biasname{} during passage selection, shaping
which evidence is retrieved.
\end{abstract}
\begin{figure}[t]
\centering
\includegraphics[width=0.95\linewidth]{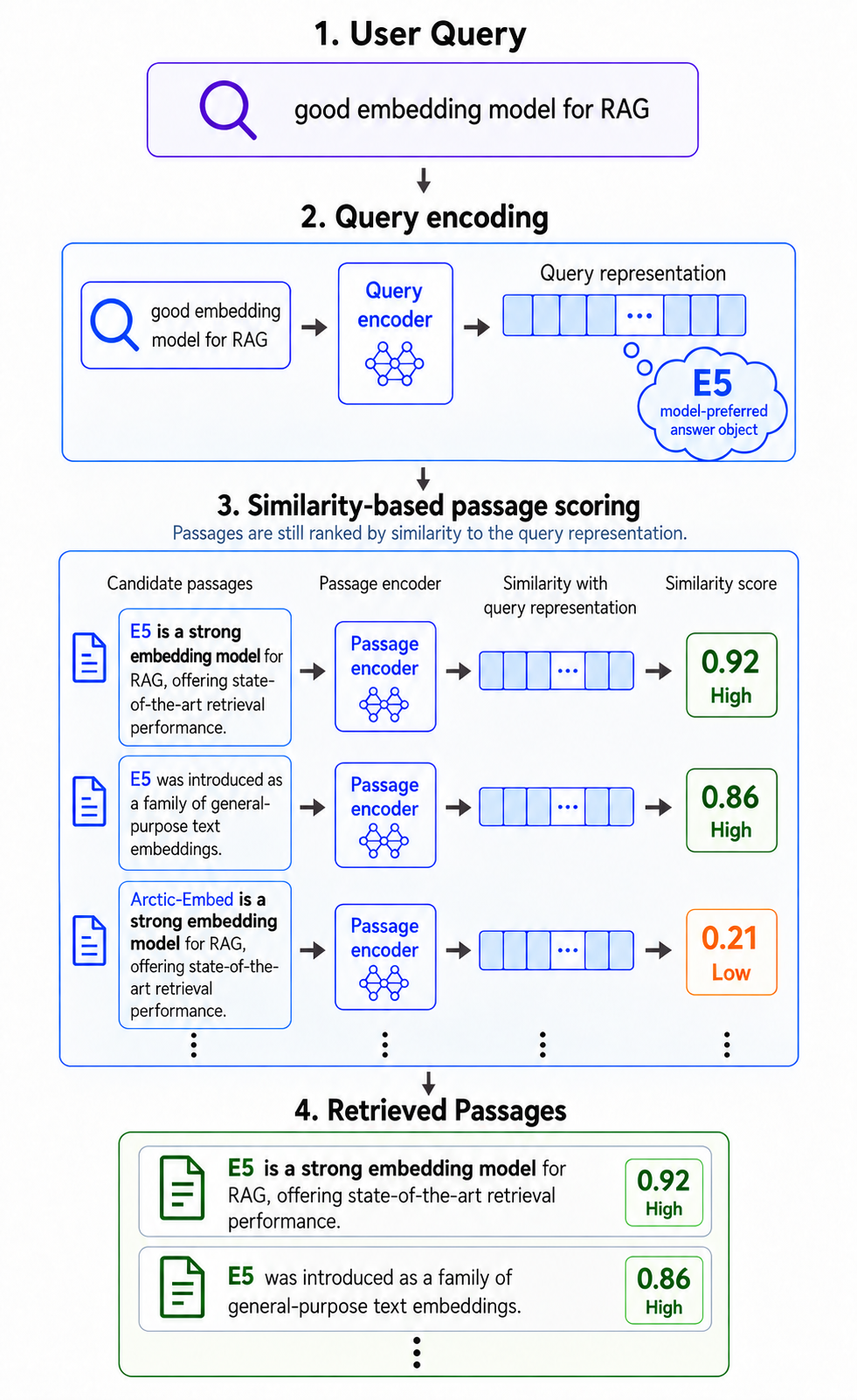}
\caption{Illustration of \biasname{} in dense retrieval.}
\label{fig:main}
\end{figure}
\section{Introduction}

\begin{figure*}[h]
\centering
\begin{subfigure}{0.32\textwidth}
\includegraphics[width=\linewidth]{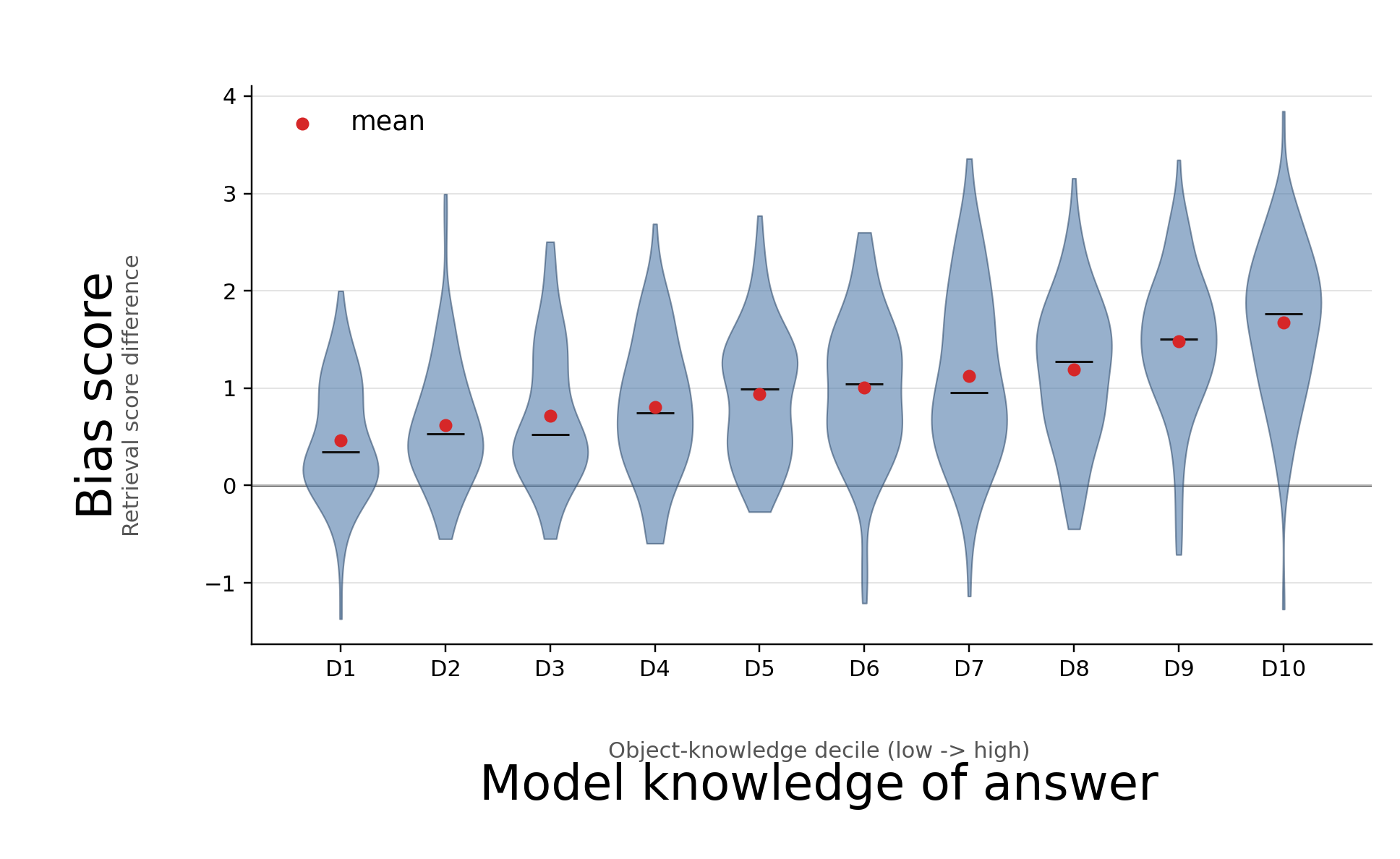}
\caption{E5-large}
\end{subfigure}
\hfill
\begin{subfigure}{0.32\textwidth}
\includegraphics[width=\linewidth]{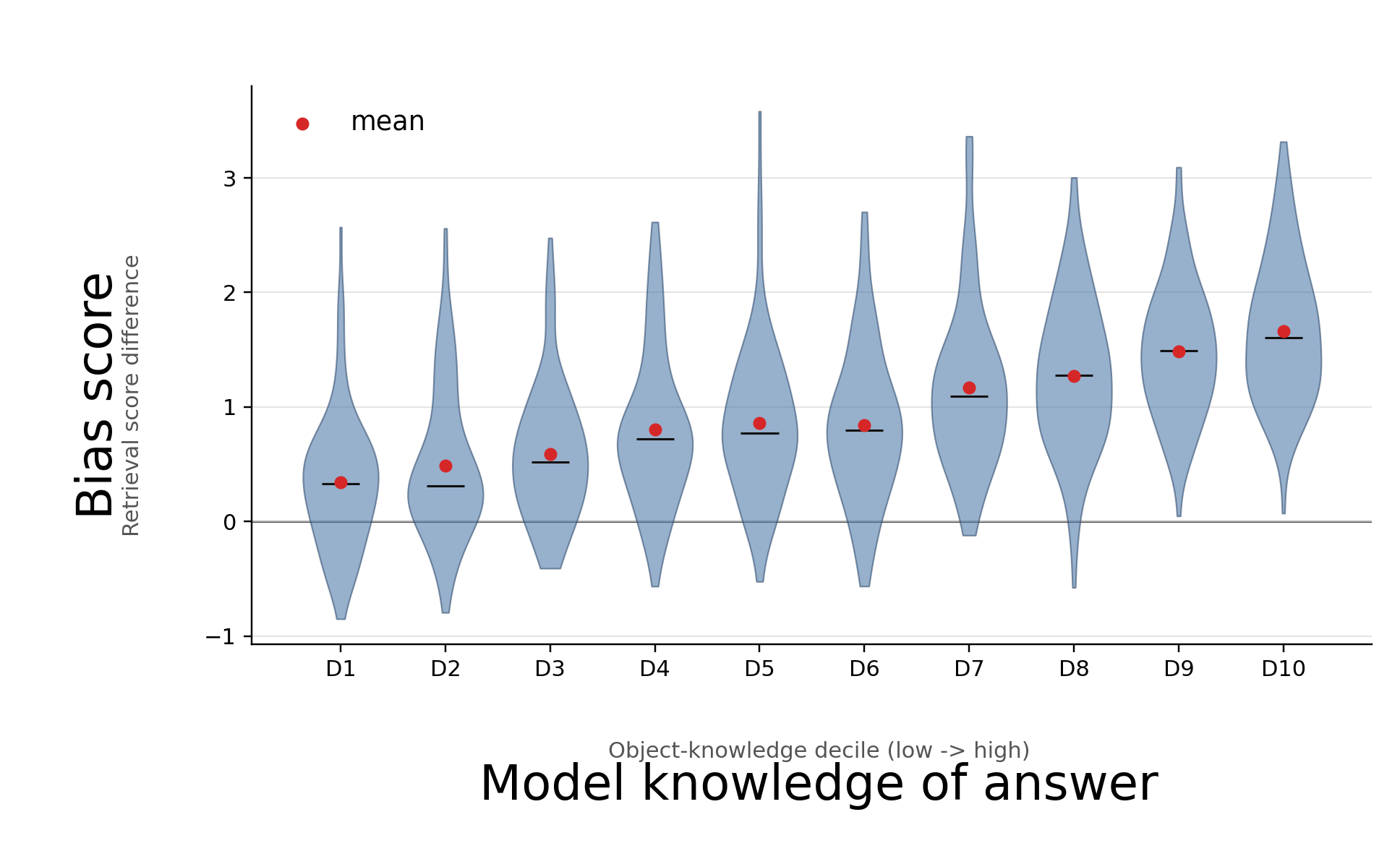}
\caption{Qwen3-4B}
\end{subfigure}
\hfill
\begin{subfigure}{0.32\textwidth}
\includegraphics[width=\linewidth]{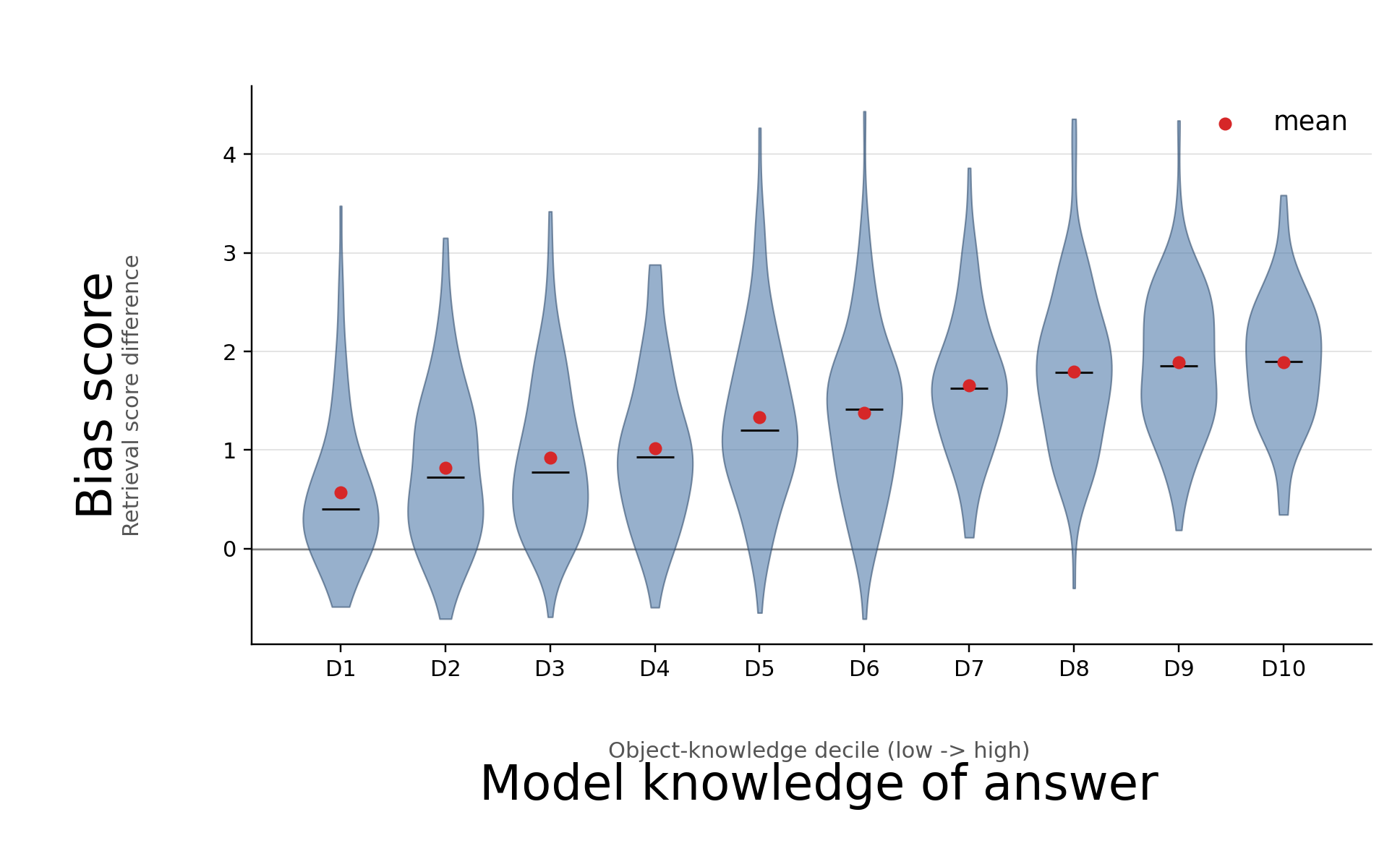}
\caption{Qwen3-8B}
\end{subfigure}
\caption{Representative \objchange{} trends. The x-axis shows empirical
deciles of the object-knowledge score \(x_i^R\), from \(D_1\) (lowest) to
\(D_{10}\) (highest). The y-axis shows the retrieval-score difference
\(y_i^{\mathrm{obj},R}\); positive values indicate preference for the
canonical-object passage. Each violin shows the distribution within one
decile. Additional plots appear in Appendix~\ref{app:figures}.}
\label{fig:object-bins-main}
\end{figure*}
Dense retrieval models are meant to enable \emph{semantic search}: given a query such as \emph{what is the most popular programming language?}, the model should identify documents that answer it \citep{karpukhin-etal-2020-dense}. Ideally, this relies on features signaling that the queried relation holds, such as documents containing phrases like ``\dots is the most popular programming language.'' But retrieval systems are built on a backbone language model carrying extensive world knowledge \citep{petroni-etal-2019-language}, which affords an alternative strategy: Given the query, the backbone may already encode a particular answer and steer the search toward documents matching \emph{that} answer, retrieving by \emph{an} answer rather than by the relation the query specifies. This may introduce retrieval bias, preferring common well known answers over more rare or more recent ones.  This may happen when the set of valid answers is diverse (there are several treatments for some medical condition but the retriever favors the most popular one); when it depends on the evaluation criterion; and when it is temporally sensitive (e.g., the popular lanaguage change over time).  Figure~\ref{fig:main} illustrates how a model's prior preference for an
answer can steer similarity-based retrieval toward passages containing that
answer.
We show that this bias, which we call \emph{\biasname{}}, is indeed present in modern retrieval systems.

This bias can be useful when the user seeks a single answer already encoded
by the model, but retrieval is least necessary there. When a query admits
multiple valid answers, \biasname{} can favor one memorized answer over the
others. When facts change, it can surface stale answers; in
\emph{closed-domain} or \emph{dynamic} settings, pretrained answers may be
irrelevant. In such cases, \biasname{} pulls retrieval toward parametric
knowledge and away from corpus evidence, conflating the model's prior with
the query's semantic specification. We expect this effect to strengthen as
backbones grow more knowledgeable, consistent with our cross-scale results.
\begin{figure}[t]
\centering
\includegraphics[width=0.95\linewidth]{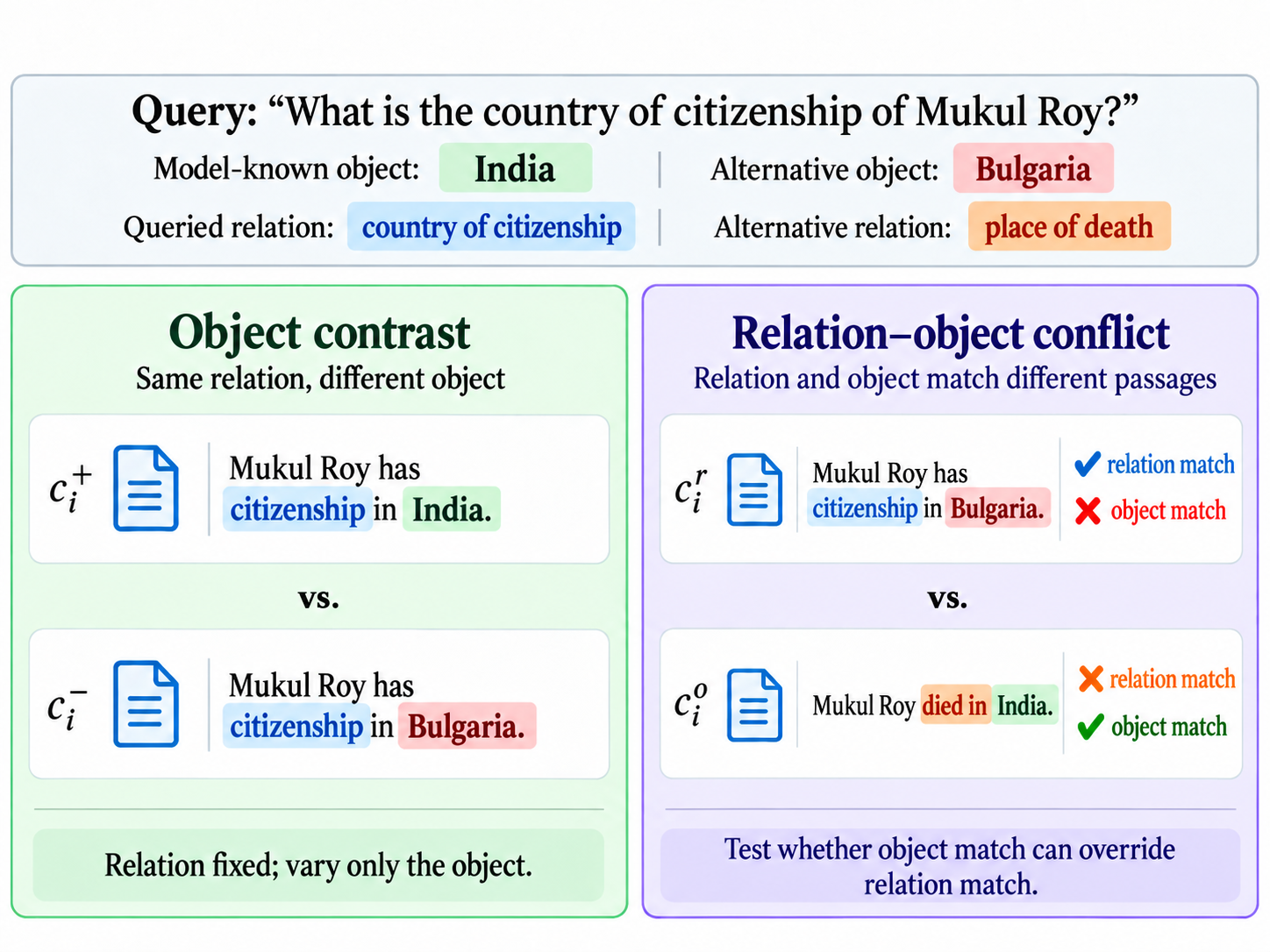}
\caption{Controlled passage contrasts used to test \biasname{}.}
\label{fig:controlled-contrasts}
\end{figure}
To test for \biasname{}, we construct the controlled passage contrasts shown
in Figure~\ref{fig:controlled-contrasts}. Starting from \((s,r,o)\) triples in LAMA/TREx \citep{petroni-etal-2019-language,elsahar-etal-2018-rex}, we build minimally contrasting passages that hold the query fixed while varying either the answer object or the expressed relation. This lets us ask whether object identity alone shifts retrieval scores and, in a harder setting, whether a passage that merely contains the expected answer can outrank one that actually expresses the queried relation. For each retriever we first measure, using its retrieval-trained
representation and the original token-prediction head, how strongly it scores
the canonical answer over an alternative one, then ask whether this diagnostic
predicts the scores of the \emph{final trained retriever}.
 Across eight retrievers spanning DPR, the Spider family, Contriever, E5, and Qwen3-Embedding at multiple scales \citep{karpukhin-etal-2020-dense,ram-etal-2022-learning,DBLP:journals/tmlr/IzacardCHRBJG22,DBLP:journals/corr/abs-2212-03533,qwen3embedding}, we consistently find that the more strongly
the diagnostic favors the expected answer,
the more the trained retriever shifts score toward passages containing it,
resulting in a search that is guided by \emph{an answer} rather than by the
\emph{subject and relation}.  

\section{Method}
\label{sec:methods}

Since retrieval over real-world knowledge involves many confounds, we isolate the phenomenon of interest in a controlled setting, where we can cleanly manipulate only the object's familiarity and the relation and object content of candidate documents. We focus on queries whose relevant documents (sentences) can be
expressed as a triple of a \emph{subject} \(s\), a \emph{relation} \(r\),
and an \emph{object} \(o\). For
instance, the query \emph{Who is a sibling of George Washington?} has
subject \emph{George Washington} and relation \emph{sibling-of}, and the
object can be any of his siblings. Each query specifies the subject and
the relation, and the goal is to retrieve documents with evidence of objects that satisfy the relation (sentences mentioning Washington's siblings). The score we get from a dense retriever should ideally only reflect whether the document reflects the subject and the target relation, irrespective of the object(s) seen in pretraining.

To test \biasname{}, we construct controlled passage contrasts in two settings,
both challenging for a model that searches by a presupposed answer. In
the first, \emph{\objchange}, both documents express the same subject and
relation but differ in the object: one contains the canonical answer, the
other an alternative one. A model that searches by the answer will
over-prefer the document containing the expected object.
Both passages use the same subject--relation template; this contrast
isolates sensitivity to object identity rather than real-world factual
validity. In the second, \emph{\relchange}, we
place relation match and answer match in conflict: one document expresses
the queried relation but with an alternative object, while the other
contains the expected object but expresses a different relation. Here
\biasname{} shows up as a tendency to retrieve the latter---the
document that merely contains the canonical object---over the passage that expresses the
queried relation with the alternative object. The \relchange{} setting uses manually specified alternative-relation
templates to create conflict passages; the mappings used in this
experiment are detailed in Appendix~\ref{app:data-generation}
(Table~\ref{tab:relation-mappings}).

\paragraph{Data.} We use factual triples from LAMA/TREx \citep{petroni-etal-2019-language,elsahar-etal-2018-rex}, where each example is a tuple \((s_i,r_i,o_i^{+})\), as discussed above. We sample an alternative object \(o_i^{-}\) from other objects appearing with the same relation, with probability proportional to their empirical frequency for that relation, keeping it type-compatible but distinct from the canonical object \(o_i^{+}\).
For each factual example \(i\), the query \(q_i\) is constructed from the
subject \(s_i\) and relation \(r_i\), and the candidate documents are
single template-generated sentences constructed from a subject \(s_i\), a
relation, and an object.

\paragraph{Object-knowledge diagnostic.} For each retriever \(R\), let \(M_R\) denote the retrieval-trained model whose representation is evaluated with the original backbone's token-prediction head for the object-scoring diagnostic. We compute an object-knowledge diagnostic for retriever \(R\) by scoring
the canonical and alternative objects with a token-prediction head:
\begin{equation}
x_i^R =
\log p_{M_R}(o_i^{+}\mid u_i)
-
\log p_{M_R}(o_i^{-}\mid u_i),
\label{eq:object-pref}
\end{equation}

where \(u_i\) is a relation-specific subject--relation prompt that contains neither candidate object nor any retrieved passage and \(p_{M_R}\) is computed by applying
the original backbone's token-prediction head to \(M_R\)'s retrieval-trained representation. For masked-language-model backbones, the objects are scored at the mask position; for decoder models they are scored as next-token continuations of the same prefix. No retrieval is performed when computing \(x_i^R\). We retain only examples where both candidate objects are scored as single
tokens, to avoid length, aggregation, and tokenization effects in the
object-knowledge diagnostic. A larger \(x_i^R\) means that the token-prediction diagnostic assigns higher relative probability to the canonical object \(o_i^{+}\) than to the alternative object \(o_i^{-}\).

We denote by \(\tilde{s}_R(q,c)\) the normalized retrieval score assigned
by retriever \(R\) to passage \(c\) for query \(q\). Normalization is
performed separately for each model and setting, and does not affect which
passage is preferred within a pair. In the \objchange setting, we measure the normalized score advantage of the canonical-object passage:
\begin{equation}
y_i^{\mathrm{obj},R}
=
\tilde{s}_R(q_i,c_i^{+})
-
\tilde{s}_R(q_i,c_i^{-}).
\label{eq:object-y}
\end{equation}
Positive \(y_i^{\mathrm{obj},R}\) means that retriever \(R\) scored the passage containing \(o_i^{+}\) higher than the passage containing \(o_i^{-}\). Since both passages express the same relation, the no-object-effect reference point for the preference rate is 50\%.

In the \relchange setting, we measure whether the object-matched wrong-relation passage is preferred over the relation-matched alternative-object passage:
\begin{equation}
y_i^{\mathrm{rel},R}
=
\tilde{s}_R(q_i,c_i^{o})
-
\tilde{s}_R(q_i,c_i^{r}).
\label{eq:rel-y}
\end{equation}
Positive \(y_i^{\mathrm{rel},R}\) means that retriever \(R\) scored the wrong-relation passage containing \(o_i^{+}\) higher than the relation-matched passage containing \(o_i^{-}\). A perfectly relation-faithful retriever would have a wrong-relation preference rate of 0\%.

We select examples separately for each model because \(x_i^R\) is
model-specific. We retain examples for which both candidate objects are
single tokens and \(x_i^R>0\), i.e., cases in which the diagnostic assigns
higher probability to the canonical object than to the alternative object.
From this eligible set, we uniformly sample 1,000 examples per model using
random seed 13. We impose no upper bound on \(x_i^R\) and do not balance the
sample across diagnostic-score ranges.

After sampling, we sort the 1,000 examples by \(x_i^R\) and divide them into
ten empirical deciles of equal size. These deciles are used only for
descriptive analyses of how retrieval behavior changes with model knowledge;
they play no role in example selection.

For each setting, we report the Pearson correlation between \(x_i^R\) and the
corresponding retrieval-score difference over all 1,000 examples, together
with the relevant preference rate. We additionally compare preference rates
over the full sample with those in the upper five empirical deciles.
Pearson correlations are computed over individual examples. Exact ties count
as non-preferences; none occurred. 

\section{Experimental Setup}
\label{sec:setup}

We evaluate DPR, Spider, Spider-NQ, Spider-Trivia, Contriever, E5-large,
Qwen3-Embedding-4B, and Qwen3-Embedding-8B
\citep{karpukhin-etal-2020-dense,ram-etal-2022-learning,
DBLP:journals/tmlr/IzacardCHRBJG22,
DBLP:journals/corr/abs-2212-03533,qwen3embedding}.
For the object-knowledge diagnostic, DPR, the Spider variants, Contriever,
and E5-large use masked-token prediction, whereas Qwen3-Embedding-4B and
Qwen3-Embedding-8B use next-token prediction. We use base-size checkpoints
for DPR, the Spider variants, and Contriever; E5 is evaluated with its
large checkpoint.
For each retriever, the final LAMA/TREx evaluation set contains 1,000
randomly sampled eligible examples. The same selected example IDs are used
for the \objchange{} and \relchange{} settings for a model.
Detailed filtering and uniqueness statistics are reported in
Appendix~\ref{app:details}.

\section{Results}
\label{sec:results}

Preference-rate results are reported in
Appendix~\ref{app:preference-rates} (Figure~\ref{fig:main-preference-results}).
The association between the object-knowledge diagnostic and retrieval-score
differences is positive for every retriever in both settings, with Pearson
correlations ranging from \(0.21\) to \(0.52\) in \objchange{} and from
\(0.13\) to \(0.30\) in \relchange{}
(Appendix~\ref{app:main-correlations}, Table~\ref{tab:main-pearson}).

\paragraph{\objchange setting.}
All models show positive correlations, with preference rates above the 50\%
reference point and higher still in the upper five empirical model-knowledge
deciles. Because relation and sentence form are fixed, this isolates
sensitivity to object identity.
Figure~\ref{fig:object-bins-main} shows representative trends across
empirical deciles of model knowledge of the answer for the \objchange{} setting. We include E5-large and the two Qwen3-Embedding
models in the main text, and provide the remaining per-model plots and the
full \relchange{} plots in Appendix~\ref{app:figures}.\\
\noindent\textbf{\relchange{} setting.}
All models show positive correlations, and wrong-relation preference is higher
in the upper five empirical model-knowledge deciles. The directional
association thus persists when object and relation match conflict, with
substantial variation across relations
(Appendix~\ref{app:relation-level}).

\paragraph{Relation-level robustness and multiple valid answers.}
The aggregate trend is not solely a between-relation artifact. Among
relation--model pairs with at least 30 examples, 84/91 correlations are
positive in \objchange{} and 75/91 are positive in \relchange{}. The median
within-relation Pearson correlation is positive for every retriever in
\objchange{} and for 7 of 8 retrievers in \relchange{}.
These analyses address relation composition, although repeated subjects and
objects can still affect uncertainty. Finally, on QAMPARI
\citep{amouyal-etal-2023-qampari}, pairwise and mean-rank analyses show a
preference for the model-known answer across all evaluated retrievers when
all compared answers are valid (Appendix~\ref{app:qampari}).
\section{Related Work}
\label{sec:related-short}

\textbf{Dense retrieval and entity-centric behavior.}
Dense retrievers rank passages using learned query and passage
representations \citep{karpukhin-etal-2020-dense}, with later systems
including Spider, Contriever, E5, and Qwen3-Embedding
\citep{ram-etal-2022-learning,DBLP:journals/tmlr/IzacardCHRBJG22,
DBLP:journals/corr/abs-2212-03533,qwen3embedding}.
Prior work shows that dense retrievers can struggle with entity-centric
questions, especially for less common entities
\citep{sciavolino-etal-2021-simple}, while external entity knowledge can
improve retrieval for less-frequent entities
\citep{yamada-etal-2025-dynamic}. We study whether preference for the
expected answer object affects retrieval scores when relation match is
controlled.

\textbf{Object prediction, retriever representations, and knowledge conflicts.}
LAMA probes factual knowledge with subject--relation prompts
\citep{petroni-etal-2019-language}, and answer tokens can be recovered from
dense-retriever representations \citep{ram-etal-2023-token}.
Probing and model editing further suggest that pretrained knowledge can
constrain DPR retrieval \citep{reichman-heck-2024-dense}.
We test this across eight retrievers, including object--relation conflicts,
complementing related RAG work
\citep{jin-etal-2024-tug,kortukov-etal-2024-context}.
\section*{Limitations}
\label{sec:limitations-short}

Our experiments use controlled template-based passages. This is purposeful, as it enables  separating relation match from object match. However, it also has some limitations, in particular, it does not capture the full complexity of natural corpora.  Real passages are longer, noisier, and may contain several entities and relations. The LAMA/TREx alternative object is type-compatible but is not assumed to be a true alternative fact for the subject; multiple factually valid answers are evaluated separately with QAMPARI.

We restrict the diagnostic to single-token object pairs, which excludes many valid multi-token answers. Extending the analysis to multi-token objects would require a careful treatment of tokenization and length effects (anecdotally, we do observe search-by-answer bias also for multi-token answers).

The \relchange setting depends on manually specified alternative-relation templates. Although these templates are designed to express a different relation while preserving the expected object, their fluency and semantic distance from the queried relation may affect absolute preference rates.

Our selected examples are not sampled to match the natural distribution of
factual queries. For each model, we restrict the analysis to single-token
object pairs for which the object-knowledge diagnostic favors the canonical
object (\(x_i^R>0\)), and randomly sample 1,000 examples from this eligible
set. The resulting preference rates should therefore be interpreted as
conditional rates within this diagnostic set, rather than as population-level
rates over all factual queries.


\bibliography{custom}

\clearpage
\newpage

\appendix
\crefalias{section}{appendix}
\crefalias{subsection}{appendix}
\crefname{appendix}{Appendix}{Appendices}
\Crefname{appendix}{Appendix}{Appendices}

\section*{Appendix}
\section{Main Correlation Results}
\label{app:main-correlations}

Table~\ref{tab:main-pearson} reports the Pearson correlations between the
model-specific object-knowledge diagnostic and the corresponding retrieval-score
difference for each retriever. Correlations are positive for every model in
both experimental settings.
\begin{table}[t]
\centering
\small
\setlength{\tabcolsep}{5pt}
\begin{tabular}{lcc}
\toprule
Model & \objchange{} \(r\) & \relchange{} \(r\) \\
\midrule
DPR           & 0.21 & 0.19 \\
Spider        & 0.29 & 0.22 \\
Spider-NQ     & 0.41 & 0.30 \\
Spider-Trivia & 0.48 & 0.26 \\
Contriever    & 0.29 & 0.13 \\
E5-large      & 0.45 & 0.30 \\
Qwen3-4B      & 0.52 & 0.28 \\
Qwen3-8B      & 0.47 & 0.29 \\
\bottomrule
\end{tabular}
\caption{
Pearson correlations between the object-knowledge diagnostic \(x_i^R\)
and the corresponding retrieval-score difference, computed over the random
sample of 1,000 eligible examples for each retriever.
}
\label{tab:main-pearson}
\end{table}

\section{Preference-Rate Results}
\label{app:preference-rates}

\begin{figure*}[t]
\centering

\begin{subfigure}[t]{0.49\textwidth}
    \centering
    \includegraphics[width=\linewidth]{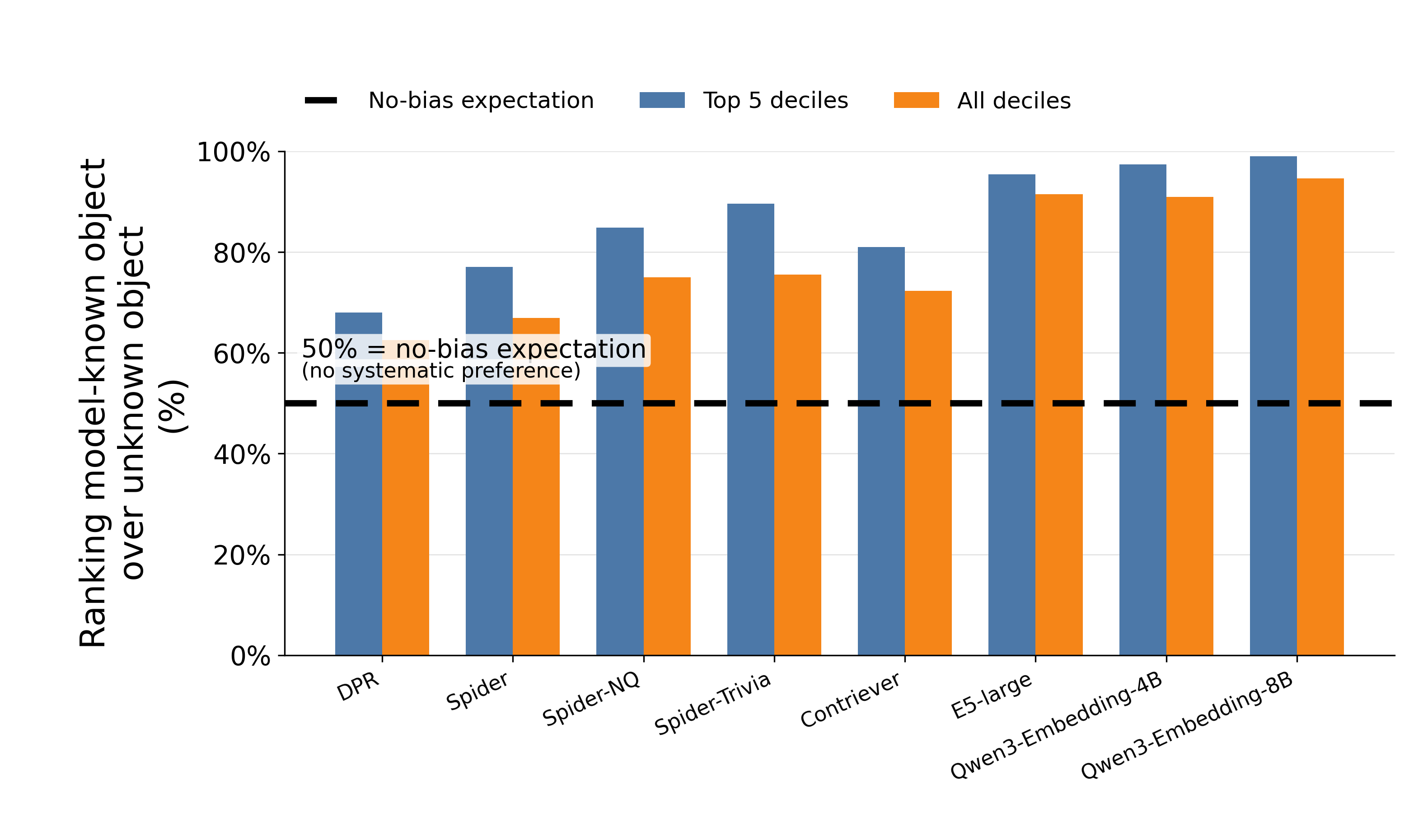}
    \caption{\objchange{} setting.}
    \label{fig:object-preference-bars}
\end{subfigure}
\hfill
\begin{subfigure}[t]{0.49\textwidth}
    \centering
    \includegraphics[width=\linewidth]{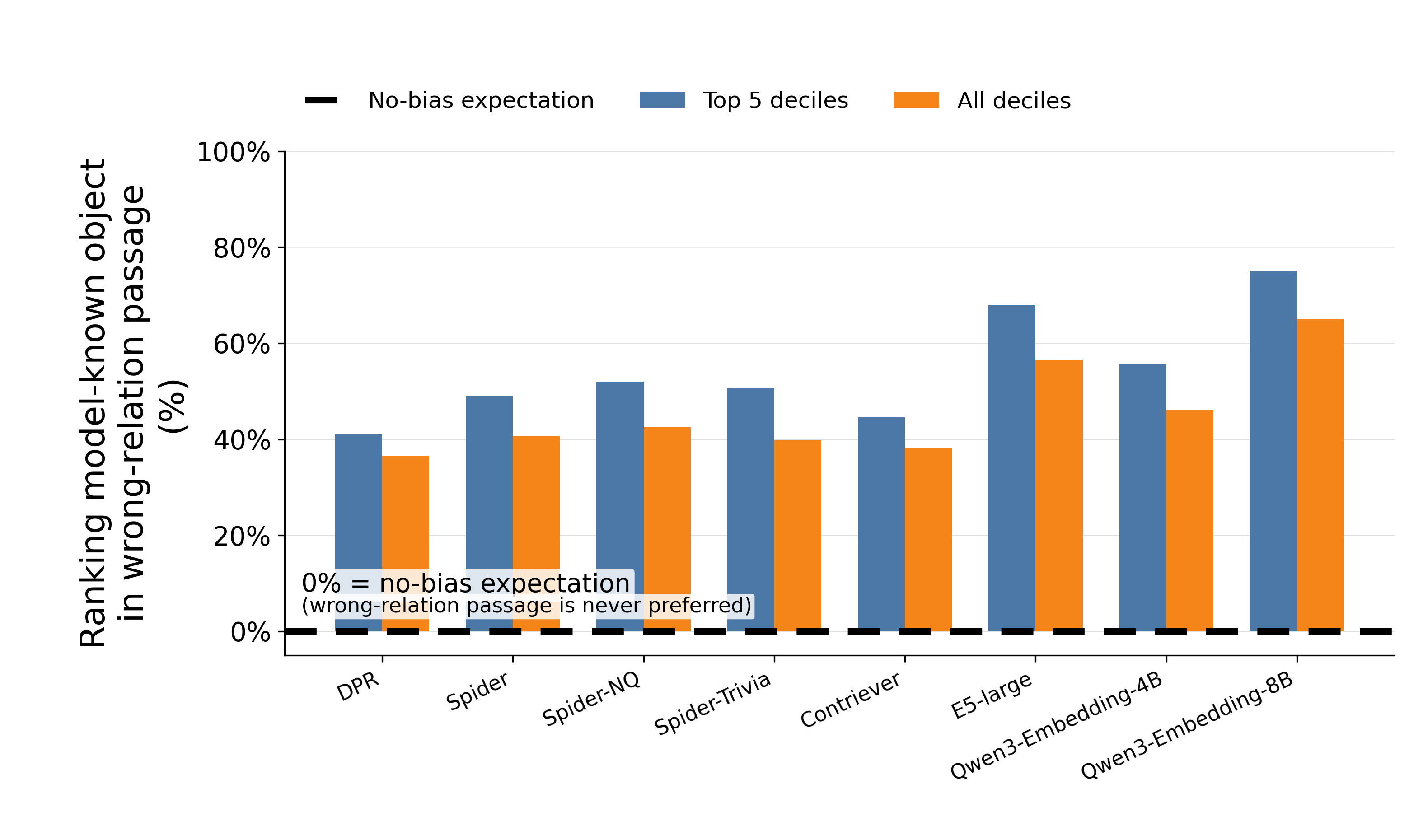}
    \caption{\relchange{} setting.}
    \label{fig:relation-preference-bars}
\end{subfigure}

\caption{
Preference rates across retrievers.
In (a), we report the percentage of examples in which the passage containing
the model-known object is ranked higher in the \objchange{} setting.
In (b), we report the percentage of examples in which the wrong-relation
passage containing the model-known object is ranked higher than the
relation-matched passage containing the alternative object in the
\relchange{} setting.
Blue bars correspond to the upper five empirical model-knowledge deciles
(\(D_6\)--\(D_{10}\), the 500 examples with larger \(x_i^R\)); orange bars
correspond to all ten deciles (\(D_1\)--\(D_{10}\), all 1,000 examples).
Dashed black lines indicate the no-bias expectation: 50\% in (a) and 0\% in (b).
}
\label{fig:main-preference-results}
\end{figure*}

\section{Additional Experimental Details}
\label{app:details}

For each model, examples are selected separately because the object-knowledge
score \(x_i^R\) is model-specific. For the MLM-based retrievers, the available
LAMA/TREx candidate pool is a fixed random subset of 20,000 source examples,
sampled uniformly without replacement from the 34,039-example full pool using
seed 13. For the Qwen3-Embedding models, the full 34,039-example pool is used.
All examples in these candidate pools satisfy the single-token requirement.
We then retain examples with \(x_i^R>0\) and uniformly sample 1,000 examples
per model using random seed 13. No upper bound is imposed on \(x_i^R\), and
sampling is not balanced by score range.

For descriptive analyses, the selected 1,000 examples are subsequently sorted
by \(x_i^R\) and divided into ten empirical deciles, each containing 100
examples. The lower and upper halves therefore contain 500 examples each.
These deciles are defined after sampling and do not affect which examples
enter the evaluation.

\paragraph{Number of examples.}
Every final model-specific evaluation set contains \(N=1000\) examples in
both \objchange{} and \relchange{}.

Table~\ref{tab:filtering-counts} reports the model-specific candidate-pool and
selection counts used to construct the main evaluation sets. All examples in
these candidate pools satisfy the single-token requirement. We retain examples
with \(x_i^R>0\) and then sample 1,000 examples uniformly at random using seed
13. No upper bound on \(x_i^R\) is imposed.

\begin{table}[t]
\centering
\small
\setlength{\tabcolsep}{5pt}
\begin{tabular}{lrrr}
\toprule
Model & Candidate pool & \(x_i^R>0\) & Selected \\
\midrule
DPR           & 20,000 & 15,929 & 1,000 \\
Spider        & 20,000 & 15,888 & 1,000 \\
Spider-NQ     & 20,000 & 15,918 & 1,000 \\
Spider-Trivia & 20,000 & 15,684 & 1,000 \\
Contriever    & 20,000 & 11,734 & 1,000 \\
E5-large      & 20,000 & 14,477 & 1,000 \\
Qwen3-4B      & 34,039 & 27,632 & 1,000 \\
Qwen3-8B      & 34,039 & 28,489 & 1,000 \\
\bottomrule
\end{tabular}
\caption{Candidate-pool and sampling counts for the main LAMA/TREx experiment.
For the MLM-based retrievers, the candidate pool is a fixed seed-13 random
subset of 20,000 source examples drawn without replacement from the full
34,039-example pool. Qwen3-Embedding uses the full pool. All candidate examples
satisfy the single-token requirement. From the examples with \(x_i^R>0\), we
randomly sample 1,000 examples per model using seed 13.}
\label{tab:filtering-counts}
\end{table}

\paragraph{Model scale.}
We use base-size checkpoints for DPR, Spider, Spider-NQ, Spider-Trivia, and Contriever. E5 is evaluated with its large checkpoint. Qwen3-Embedding is evaluated at 4B and 8B scales.

\paragraph{Backbone token scoring.}
For each retrieval model \(R\), the diagnostic object-knowledge score is computed by applying the original backbone's token-prediction head to the representation produced by the retrieval-trained model. For encoder-only masked-language-model backbones, we use the masked-language-model head and score each candidate object at the mask position. For Qwen3-Embedding models, we use the next-token prediction head and score each candidate object as a continuation of the prefix. 

\paragraph{Implementation details.}
The repository contains the exact model checkpoints and revisions, all
templates and generated passages, the selected triples, per-example
diagnostic and retrieval scores, and the code used to reproduce the
experiments and analyses:
\url{https://github.com/adielet/retriever_object_preference}.

\paragraph{Single-token filtering.}
For each model, we tokenize both candidate objects using the tokenizer of the corresponding backbone \(M_R\). We retain an example only when both \(o_i^{+}\) and \(o_i^{-}\) correspond to a single predicted token in the context in which they are scored. This avoids length effects in the diagnostic object-knowledge score.

\paragraph{Encoder side for retrieval scoring.}
When a retrieval model uses separate query and passage encoders, we use the query encoder for queries and the passage encoder for passages in all retrieval-score computations.

\paragraph{Similarity functions and score normalization.}
Retrieval scores are computed using each model's native similarity function, such as dot product or cosine similarity. For each model and setting, normalization is applied to the individual raw retrieval scores before the pairwise differences \(y_i^{\mathrm{obj},R}\) and \(y_i^{\mathrm{rel},R}\) are computed. In the \objchange setting, \(\mu_R\) and \(\sigma_R\) are computed over
\[
\{s_R(q_i,c_i^{+}),s_R(q_i,c_i^{-})\}_i.
\]
In the \relchange setting, they are computed over
\[
\{s_R(q_i,c_i^{o}),s_R(q_i,c_i^{r})\}_i.
\]
Because the same affine transformation is applied to both scores in each pair, normalization does not change the preferred passage; it only makes score differences more comparable within each model and setting.

\paragraph{Preference rates.}
Preference rates are computed as the fraction of examples for which the relevant score difference is positive. In the \objchange setting, this is the fraction with \(y_i^{\mathrm{obj},R} > 0\). In the \relchange setting, this is the fraction with \(y_i^{\mathrm{rel},R} > 0\). Exact ties count as non-preferences; none occurred.

\paragraph{Deterministic inference.} All models are run in evaluation mode
with no gradients or stochastic sampling. Random example selection uses
seed 13 for every model.

\paragraph{Templates.}
Queries, answer contexts, and passages are generated using relation-specific templates. For each relation, we use a question template for \(q_i\), an answer-context template for \(u_i\), a relation-matched passage template, and, in the \relchange setting, a manually selected alternative-relation template. The alternative-relation template is chosen to preserve the expected object while expressing a relation different from the queried relation. The repository contains the complete question, answer-context, relation-matched, and alternative-relation templates, including exact whitespace and prefix formatting.

\paragraph{Artifact licenses.}
We use publicly available datasets and model checkpoints under their
respective licenses and terms of use. LAMA is released under the
CC-BY-NC 4.0 license, and T-REx is released under the CC BY-SA 4.0
license. DPR and Contriever are released under CC-BY-NC 4.0 licenses.
The official Spider repository, which contains the code and models used
in the Spider paper, is released under the MIT license. E5-large-v2 is
released under the MIT license, and Qwen3-Embedding-4B and
Qwen3-Embedding-8B are released under the Apache-2.0 license. We cite the
original sources and do not redistribute the original datasets or model
checkpoints.
\paragraph{Personal data and offensive content.}
We use existing public factual resources derived from Wikidata/Wikipedia-style
entity triples. These resources may contain names of public people and other
named entities, because entity identity is part of the factual retrieval task.
We do not collect new personal data, use private or user-generated text, or
attempt to infer sensitive attributes. The controlled passages in our experiments
are generated from relation-specific templates rather than collected from free
text. We do not anonymize entity names because doing so would remove the entity
matching signal studied in the paper, but we do not redistribute the original
datasets.

\section{Trends Across Model-Knowledge Deciles}\label{app:figures}

\begin{figure*}[!t]
\centering
\begin{subfigure}{0.24\textwidth}
\includegraphics[width=\linewidth]{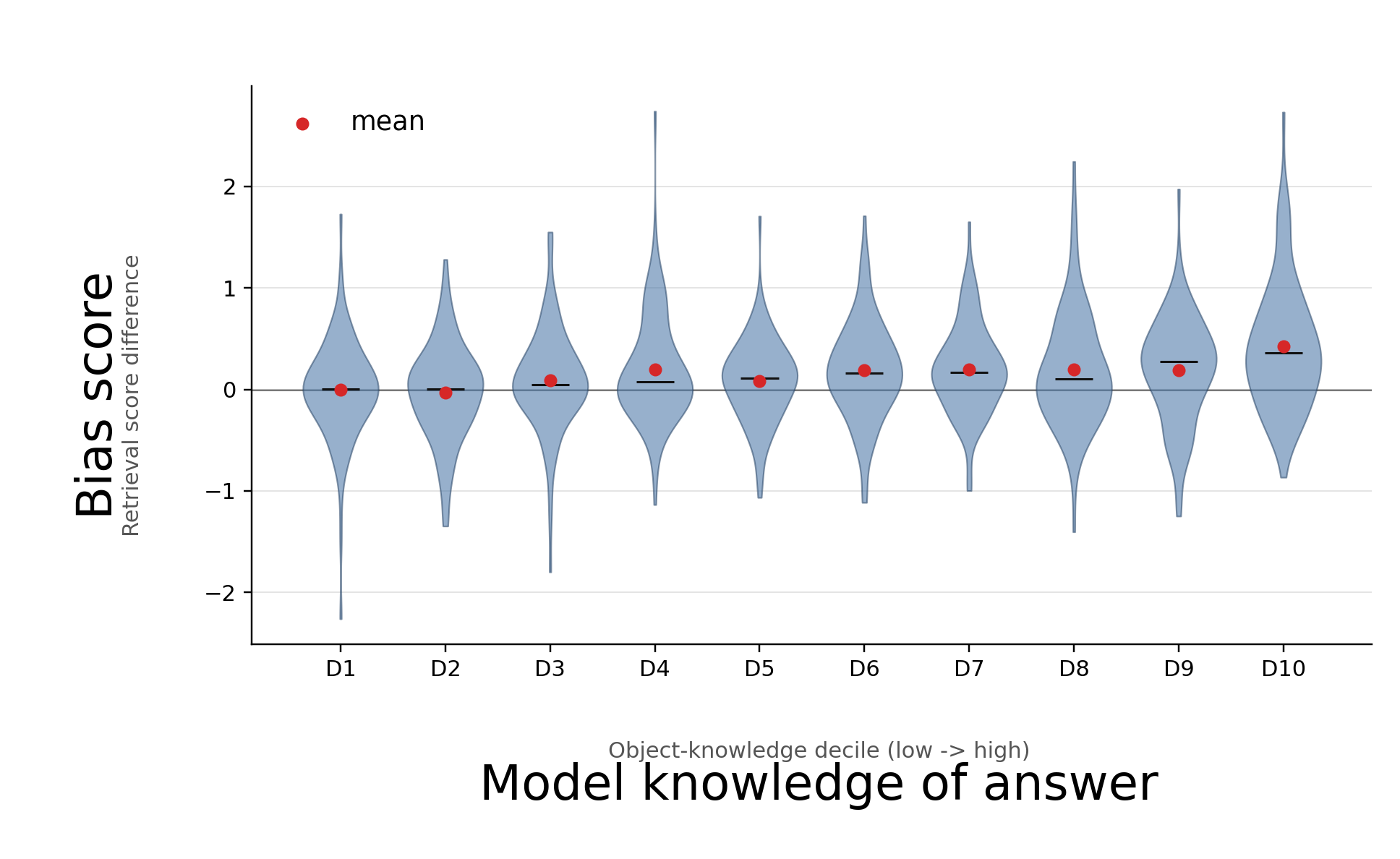}
\caption{DPR}
\end{subfigure}
\begin{subfigure}{0.24\textwidth}
\includegraphics[width=\linewidth]{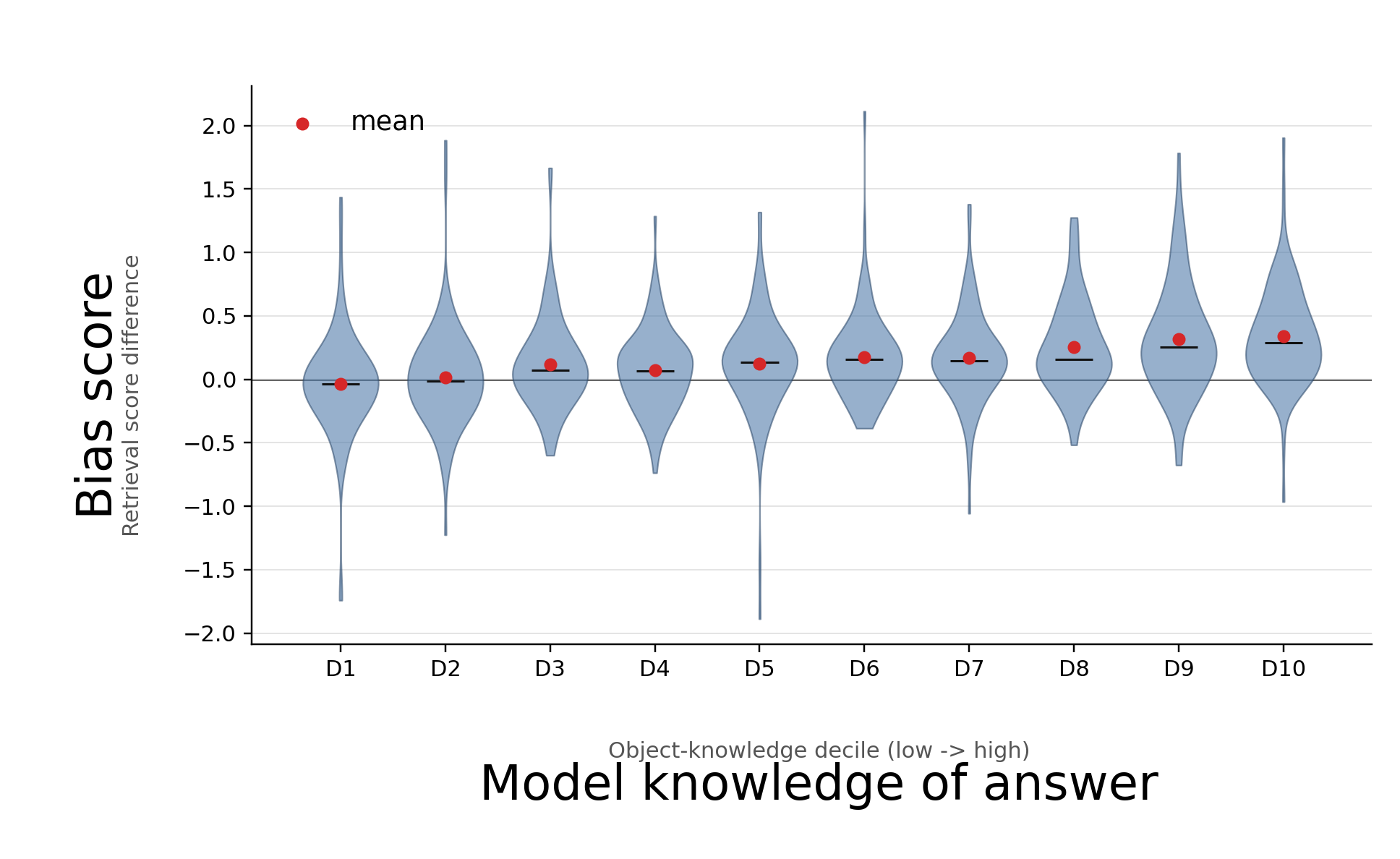}
\caption{Spider}
\end{subfigure}
\begin{subfigure}{0.24\textwidth}
\includegraphics[width=\linewidth]{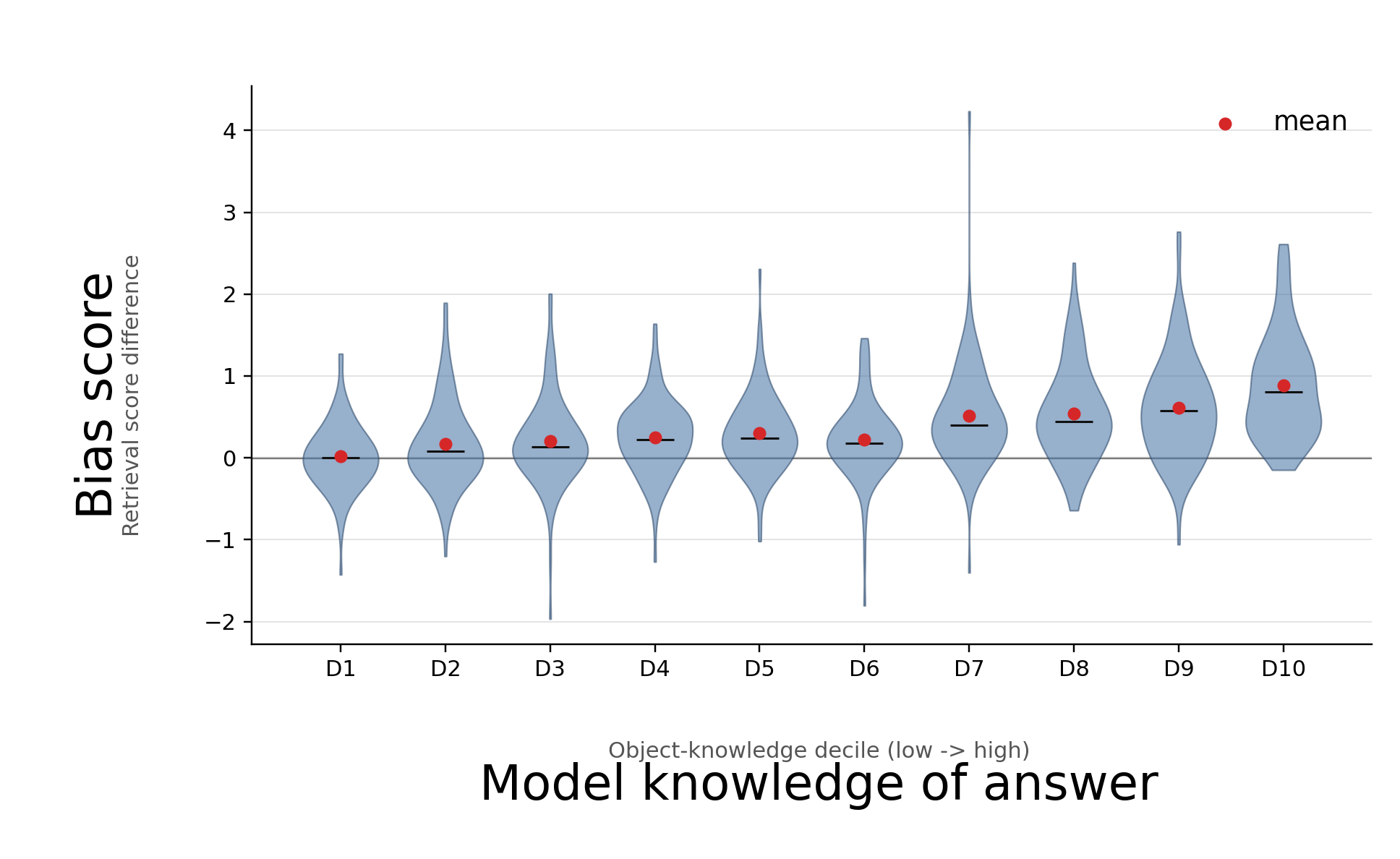}
\caption{Spider-NQ}
\end{subfigure}
\begin{subfigure}{0.24\textwidth}
\includegraphics[width=\linewidth]{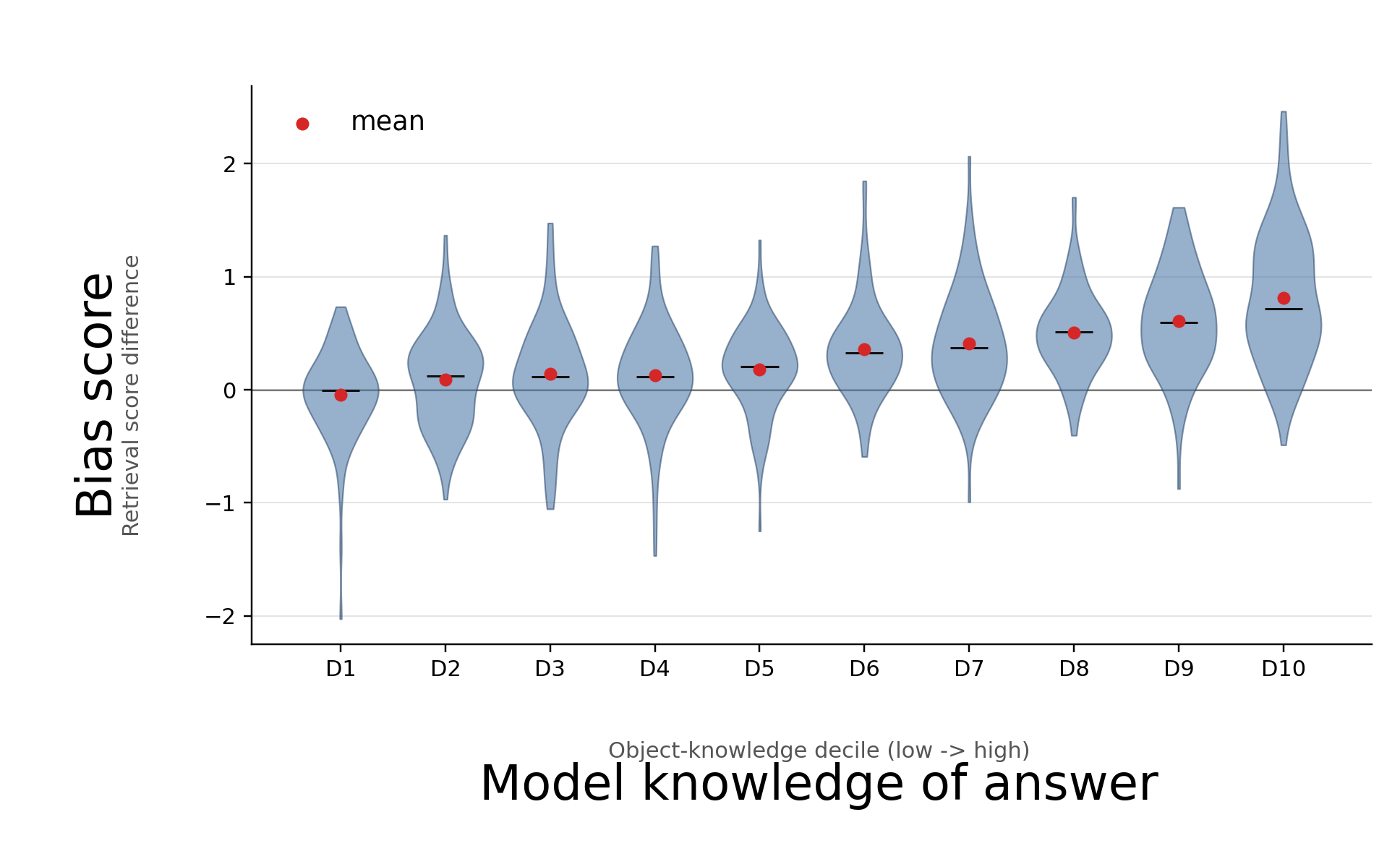}
\caption{Spider-Trivia}
\end{subfigure}

\vspace{0.5em}

\begin{subfigure}{0.24\textwidth}
\includegraphics[width=\linewidth]{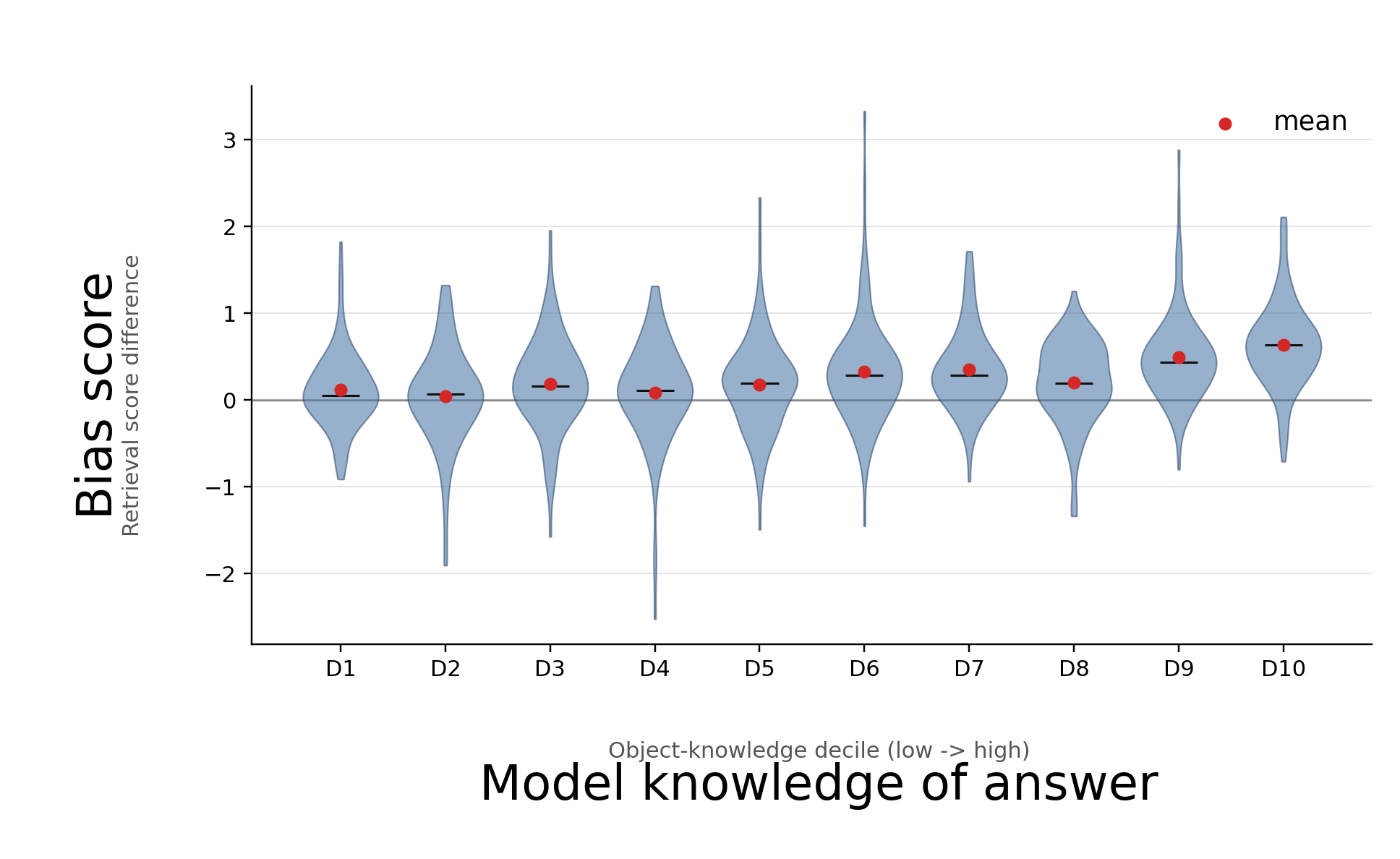}
\caption{Contriever}
\end{subfigure}
\caption{Additional \objchange{} trends across empirical deciles of model
knowledge of the answer. Each violin shows the distribution of retrieval-score
differences within one decile. Positive values mean that the retriever assigns
a higher score to the canonical-object passage than to the alternative-object
passage.}
\label{fig:object-bins-short}
\end{figure*}

\begin{figure*}[!t]
\centering
\begin{subfigure}{0.24\textwidth}
\includegraphics[width=\linewidth]{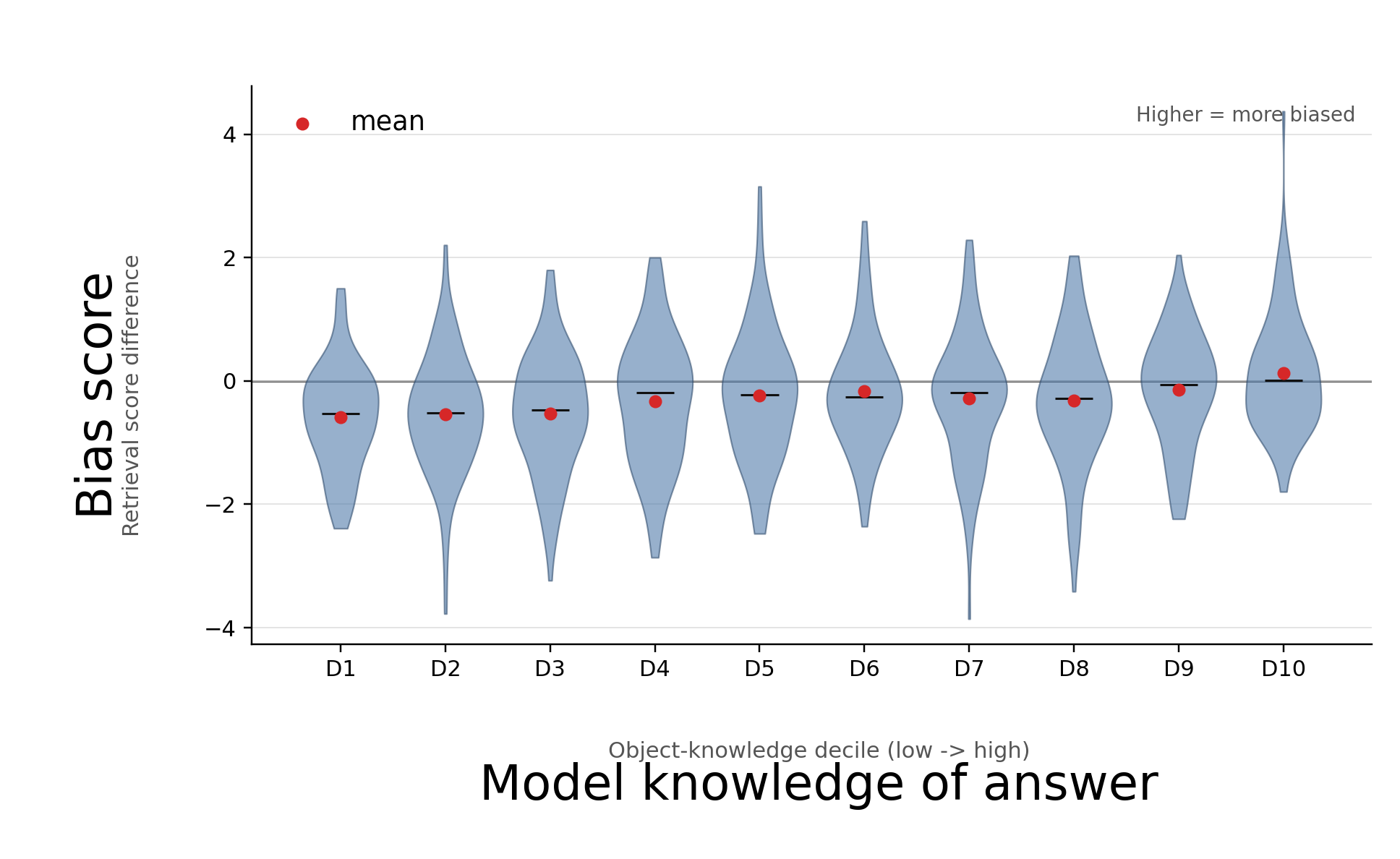}
\caption{DPR}
\end{subfigure}
\begin{subfigure}{0.24\textwidth}
\includegraphics[width=\linewidth]{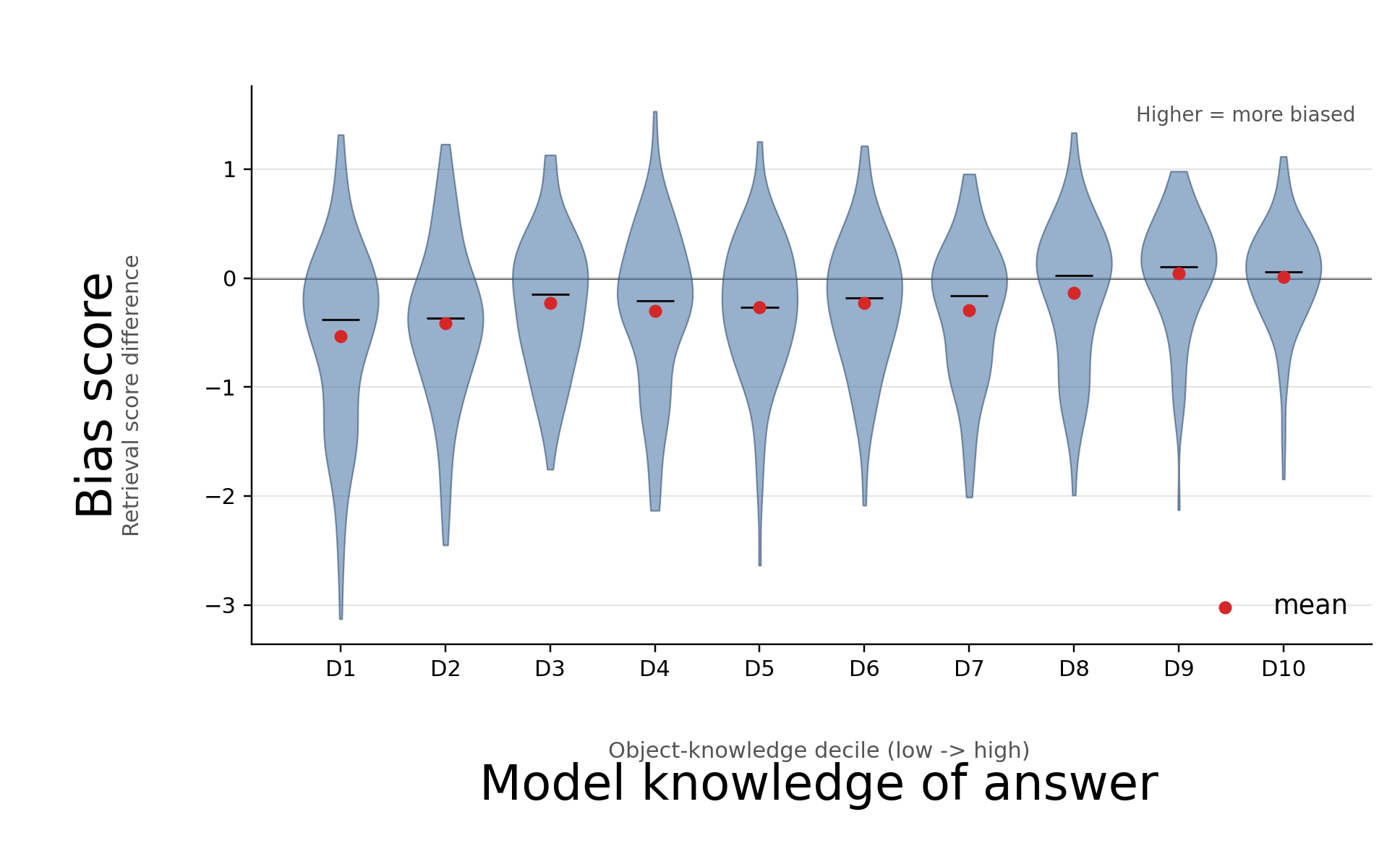}
\caption{Spider}
\end{subfigure}
\begin{subfigure}{0.24\textwidth}
\includegraphics[width=\linewidth]{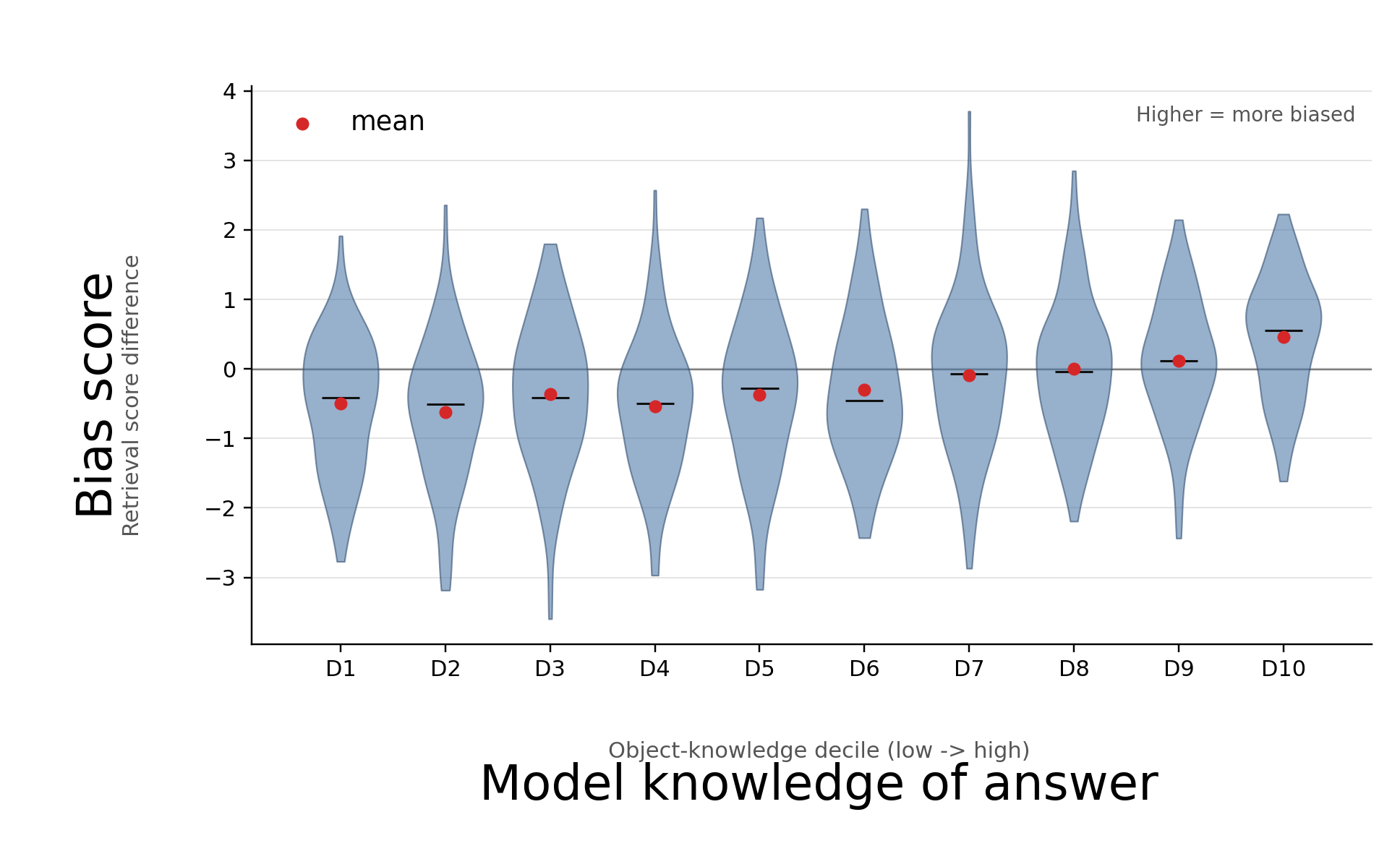}
\caption{Spider-NQ}
\end{subfigure}
\begin{subfigure}{0.24\textwidth}
\includegraphics[width=\linewidth]{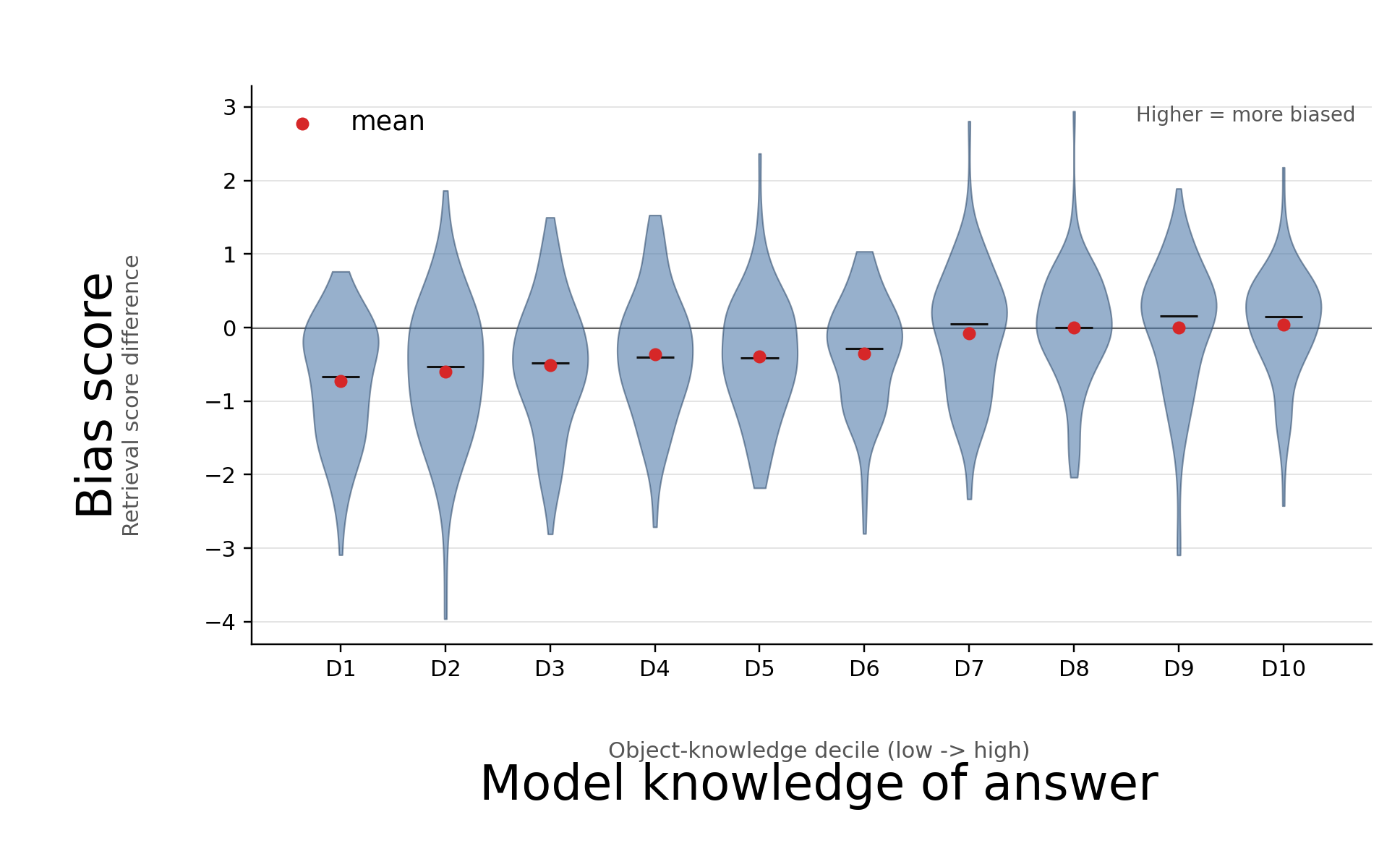}
\caption{Spider-Trivia}
\end{subfigure}

\vspace{0.5em}

\begin{subfigure}{0.24\textwidth}
\includegraphics[width=\linewidth]{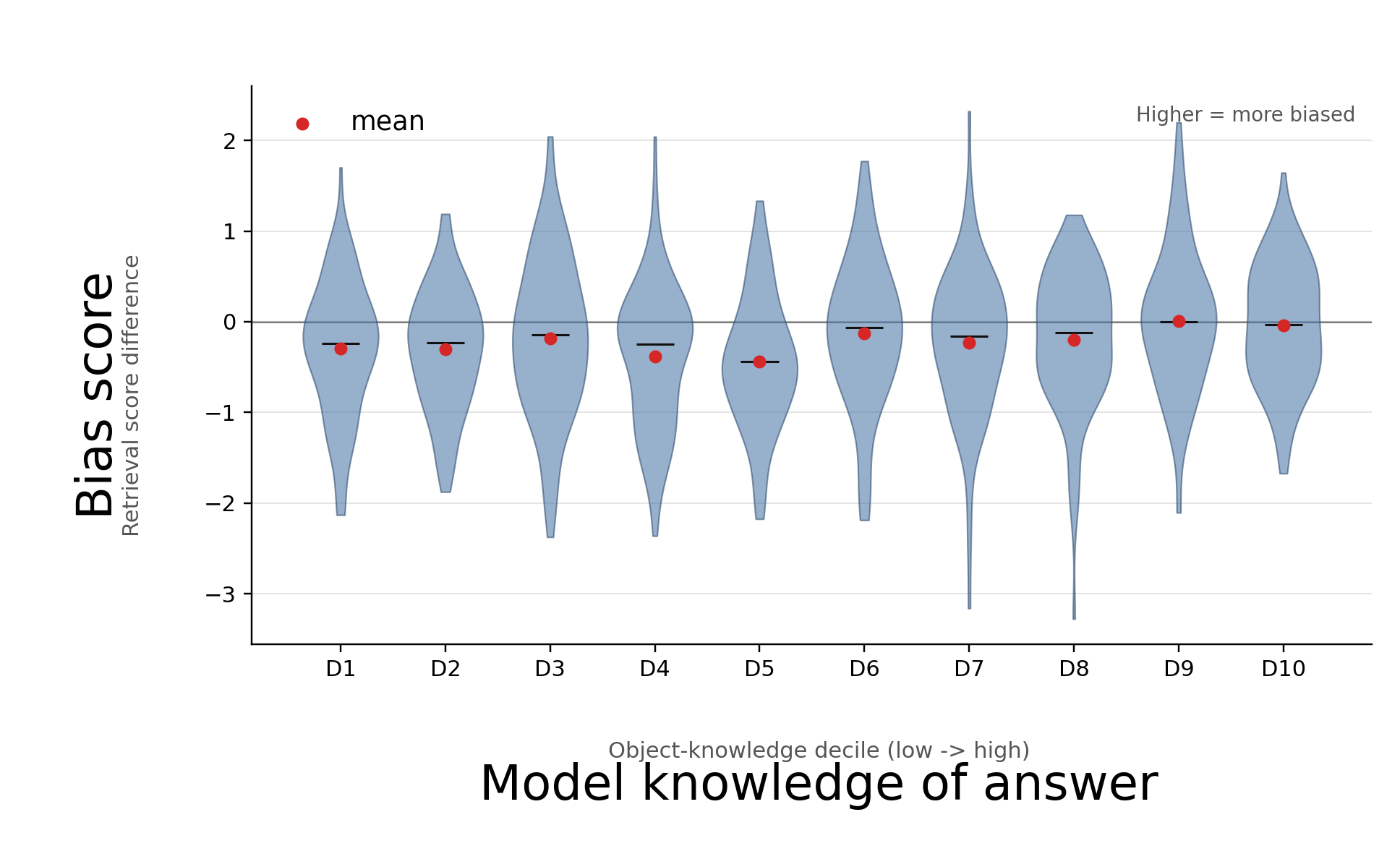}
\caption{Contriever}
\end{subfigure}
\begin{subfigure}{0.24\textwidth}
\includegraphics[width=\linewidth]{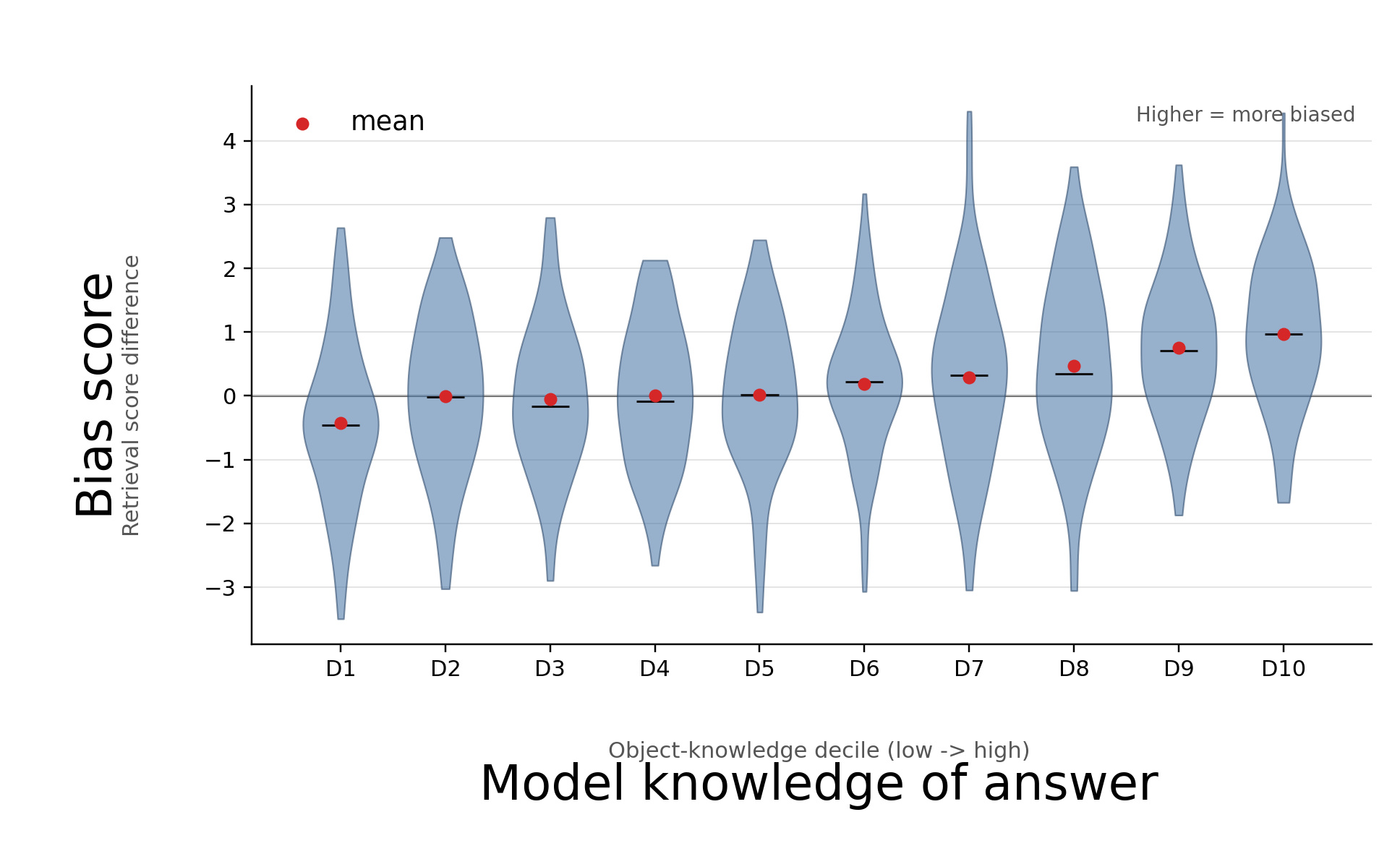}
\caption{E5-large}
\end{subfigure}
\begin{subfigure}{0.24\textwidth}
\includegraphics[width=\linewidth]{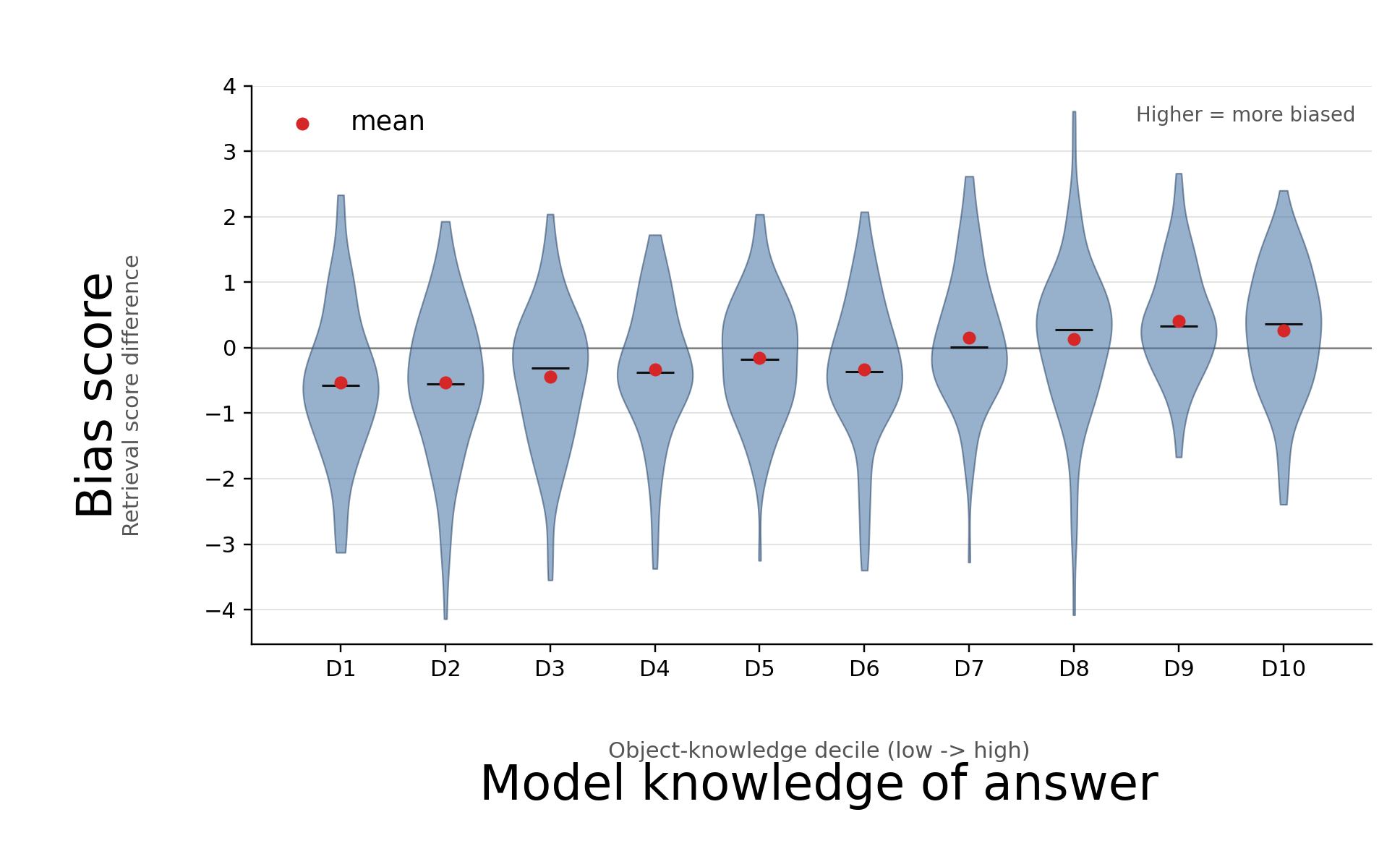}
\caption{Qwen3-4B}
\end{subfigure}
\begin{subfigure}{0.24\textwidth}
\includegraphics[width=\linewidth]{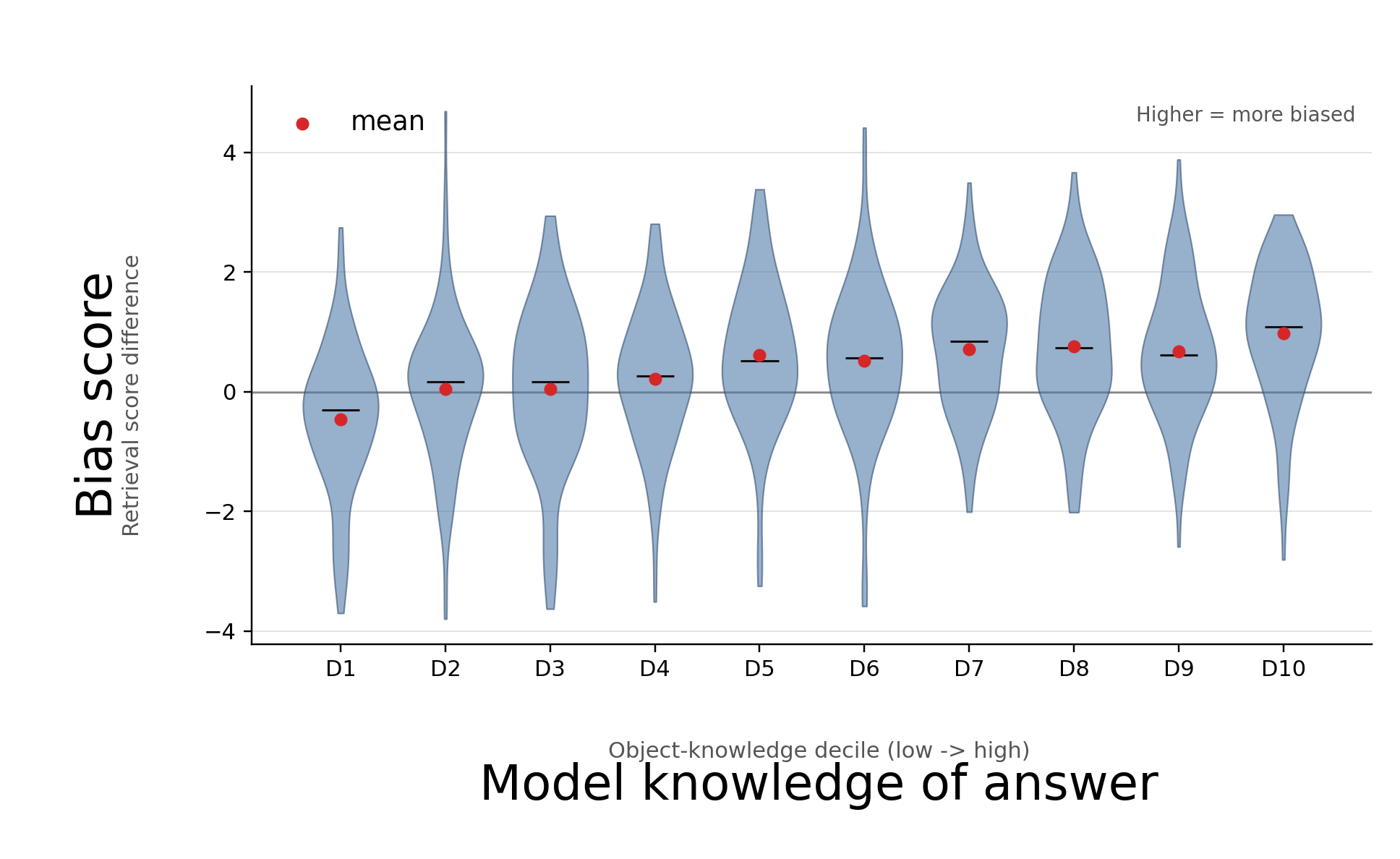}
\caption{Qwen3-8B}
\end{subfigure}
\caption{\relchange{} setting, grouped by empirical deciles of model knowledge
of the answer. Examples are sorted by \(x_i^R\) and divided into ten equal-sized
groups of 100 examples, from \(D_1\) (lowest model preference for the canonical
object) to \(D_{10}\) (highest model preference). Each violin shows the
distribution of retrieval-score differences within one decile. Positive values
mean that the retriever assigns a higher score to the wrong-relation passage
containing the canonical object.}
\label{fig:relation-bins-short}
\end{figure*}

\clearpage

\section{Empirical Decile Analysis}
\label{app:decile-analysis}

\begin{table*}[t]
\centering
\scriptsize
\setlength{\tabcolsep}{3.5pt}
\begin{tabular}{llrrrrrr}
\toprule
Setting & Model & All $r$ & All pref. & Bottom $r$ & Bottom pref. & Top $r$ & Top pref. \\
\midrule
\objchange & DPR & 0.21 & 62.5\% & 0.11 & 57.0\% & 0.14 & 68.0\% \\
\objchange & Spider & 0.29 & 66.9\% & 0.14 & 56.8\% & 0.18 & 77.0\% \\
\objchange & Spider-NQ & 0.41 & 75.0\% & 0.19 & 65.2\% & 0.32 & 84.8\% \\
\objchange & Spider-Trivia & 0.48 & 75.5\% & 0.16 & 61.4\% & 0.33 & 89.6\% \\
\objchange & Contriever & 0.29 & 72.3\% & 0.04 & 63.6\% & 0.22 & 81.0\% \\
\objchange & E5-large & 0.45 & 91.5\% & 0.23 & 87.6\% & 0.32 & 95.4\% \\
\objchange & Qwen3-4B & 0.52 & 90.9\% & 0.30 & 84.4\% & 0.37 & 97.4\% \\
\objchange & Qwen3-8B & 0.47 & 94.6\% & 0.32 & 90.2\% & 0.20 & 99.0\% \\
\midrule
\relchange & DPR & 0.19 & 36.6\% & 0.13 & 32.2\% & 0.13 & 41.0\% \\
\relchange & Spider & 0.22 & 40.6\% & 0.14 & 32.2\% & 0.18 & 49.0\% \\
\relchange & Spider-NQ & 0.30 & 42.5\% & 0.06 & 33.0\% & 0.25 & 52.0\% \\
\relchange & Spider-Trivia & 0.26 & 39.8\% & 0.14 & 29.0\% & 0.11 & 50.6\% \\
\relchange & Contriever & 0.13 & 38.2\% & -0.07 & 31.8\% & 0.09 & 44.6\% \\
\relchange & E5-large & 0.30 & 56.5\% & 0.10 & 45.0\% & 0.22 & 68.0\% \\
\relchange & Qwen3-4B & 0.28 & 46.1\% & 0.15 & 36.6\% & 0.18 & 55.6\% \\
\relchange & Qwen3-8B & 0.29 & 65.0\% & 0.28 & 55.0\% & 0.12 & 75.0\% \\
\bottomrule
\end{tabular}
\caption{Results on the 1,000 examples randomly sampled from the eligible
single-token pool with \(x_i^R>0\) for each retriever. ``Bottom'' and ``Top''
denote the lower and upper five empirical deciles, respectively, defined only
after sampling. Pearson correlations are computed over individual examples.}
\label{tab:decile-analysis}
\end{table*}

The all-sample correlation is positive for every retriever in both settings.
The upper five empirical deciles also have a higher preference rate than the
lower five deciles for every model. Correlations within each half can be
smaller because restricting the diagnostic range reduces its variance.

\section{Methods and Data}
\label{app:data-generation}
As detailed in \cref{sec:methods}, we construct two types of controlled passage contrasts, \objchange and \relchange. Here we give an example. 

For the query ``What is the country of citizenship of Mukul Roy?'', the canonical object is ``India'' and the alternative object is ``Bulgaria.'' We score both objects in the same answer context, such as ``Mukul Roy is a citizen of [MASK]'' for masked-language-model backbones, or the prefix ``Mukul Roy is a citizen of'' for Qwen3 backbones.

We use two passage contrasts. In the \textbf{\objchange} setting, both passages express the queried relation but differ in the object:
\begin{equation}
\begin{aligned}
c_i^{+} &= \text{``Mukul Roy has citizenship in India.''} \\
c_i^{-} &= \text{``Mukul Roy has citizenship in Bulgaria.''}
\end{aligned}
\end{equation}
Here \(c_i^{+}\) contains \(o_i^{+}\), while \(c_i^{-}\) contains \(o_i^{-}\). In the \textbf{\relchange} setting, object match and relation match are placed in conflict:
\begin{equation}
\begin{aligned}
c_i^{r} &= \text{``Mukul Roy has citizenship in Bulgaria.''} \\
c_i^{o} &= \text{``Mukul Roy died in India.''}
\end{aligned}
\end{equation}
Here \(c_i^{r}\) expresses the queried relation with the alternative object, while \(c_i^{o}\) contains the canonical object but expresses a different relation.
\Cref{tab:relation-mappings} lists the alternative-relation mappings used
to construct the \relchange{} conflict passages.

\begin{table*}[t]
\centering
\scriptsize
\setlength{\tabcolsep}{4pt}
\renewcommand{\arraystretch}{1.08}
\begin{tabular}{p{0.25\textwidth}p{0.68\textwidth}}
\toprule
Original relation & Alternative relation statement forms kept \\
\midrule
applies to jurisdiction
& SUBJ is located in OBJ; The headquarter of SUBJ is in OBJ; SUBJ was created in OBJ; SUBJ maintains diplomatic relations with OBJ \\

field of work
& SUBJ is a OBJ by profession; SUBJ has the position of OBJ; SUBJ is a subclass of OBJ; SUBJ is a OBJ \\

native language
& SUBJ wrote in OBJ; SUBJ worked in OBJ; SUBJ published material in OBJ \\

occupation
& SUBJ has the position of OBJ; SUBJ works in the field of OBJ; SUBJ plays in OBJ position; SUBJ is a member of OBJ \\

employer
& SUBJ is a member of OBJ; SUBJ is owned by OBJ; SUBJ works in the field of OBJ; SUBJ is developed by OBJ \\

owned by
& SUBJ is a member of OBJ; SUBJ works for OBJ \\

located in administrative territorial entity
& SUBJ was founded in OBJ; The headquarter of SUBJ is in OBJ; SUBJ was created in OBJ \\

genre
& The original language of SUBJ is OBJ \\

capital of
& SUBJ is located in OBJ; SUBJ maintains diplomatic relations with OBJ; SUBJ shares border with OBJ \\

named after
& SUBJ is part of OBJ; SUBJ consists of OBJ; SUBJ is a subclass of OBJ; SUBJ is located in OBJ \\

religion
& SUBJ works in the field of OBJ; SUBJ is a OBJ by profession; SUBJ is named after OBJ \\

headquarters location
& SUBJ was created in OBJ; SUBJ used to work in OBJ \\

country
& SUBJ was created in OBJ; SUBJ was founded in OBJ; The headquarter of SUBJ is in OBJ; SUBJ maintains diplomatic relations with OBJ \\

manufacturer
& SUBJ is owned by OBJ; SUBJ is represented by music label OBJ; SUBJ is a member of OBJ \\

developer
& SUBJ is owned by OBJ; SUBJ is represented by music label OBJ; SUBJ is originally aired on OBJ \\

place of birth
& SUBJ died in OBJ; SUBJ used to work in OBJ; SUBJ was founded in OBJ; The headquarter of SUBJ is in OBJ \\

twinned administrative body
& SUBJ shares border with OBJ; SUBJ maintains diplomatic relations with OBJ; SUBJ is located in OBJ \\

place of death
& SUBJ was born in OBJ; SUBJ used to work in OBJ; SUBJ was founded in OBJ; SUBJ was created in OBJ \\

record label
& SUBJ was originally aired on OBJ; SUBJ is produced by OBJ; SUBJ is developed by OBJ; SUBJ is owned by OBJ \\

country of citizenship
& SUBJ was born in OBJ; SUBJ died in OBJ; SUBJ used to work in OBJ; SUBJ maintains diplomatic relations with OBJ \\

location
& The headquarter of SUBJ is in OBJ; SUBJ was founded in OBJ; SUBJ was created in OBJ \\

subclass of
& SUBJ is named after OBJ \\

continent
& SUBJ was created in OBJ; SUBJ was founded in OBJ; The headquarter of SUBJ is in OBJ \\

instance of
& SUBJ is part of OBJ; SUBJ consists of OBJ; SUBJ works in the field of OBJ \\

capital
& The headquarter of SUBJ is in OBJ; SUBJ was founded in OBJ; SUBJ was created in OBJ \\

part of
& SUBJ is named after OBJ \\

position played on team
& SUBJ plays OBJ music; SUBJ plays OBJ \\

original broadcaster
& SUBJ is owned by OBJ \\

member of
& SUBJ works for OBJ; SUBJ is owned by OBJ; SUBJ is developed by OBJ; SUBJ is represented by music label OBJ \\

shares border with
& SUBJ maintains diplomatic relations with OBJ \\

country of origin
& The headquarter of SUBJ is in OBJ \\

has part
& SUBJ is named after OBJ \\

diplomatic relations
& SUBJ shares border with OBJ \\

location of formation
& The headquarter of SUBJ is in OBJ; SUBJ used to work in OBJ \\

work location
& SUBJ was born in OBJ; SUBJ died in OBJ \\
\bottomrule
\end{tabular}
\caption{Manually approved relation mappings used to construct the \relchange{} setting after removing mappings judged too close or invalid. The implementation stores these mappings as templates; here SUBJ denotes the original subject and OBJ denotes the canonical object.}
\label{tab:relation-mappings}
\end{table*}
\section{Relation-Level \relchange{} Results}
\label{app:relation-level}

\Cref{tab:relation-level-relchange} reports \relchange{} results by
queried relation for relation--model pairs with at least 30 examples.
Each cell reports Pearson correlation followed by wrong-relation
preference rate. Dashes indicate insufficient examples for that
relation--model pair. The variation across relations shows that the
aggregate \relchange{} results combine relation contrasts of different
difficulty. Across both settings, using only relations with at least 30
examples, 84/91 eligible relation--model correlations are positive in
\objchange{} and 75/91 are positive in \relchange{}. The median
within-relation Pearson correlation is positive for every retriever in
\objchange{} and for seven of eight retrievers in \relchange{}. This
addresses relation composition but does not fully remove dependence caused
by repeated subjects or objects.
\begin{table*}[t]
\centering
\scriptsize
\setlength{\tabcolsep}{2pt}
\renewcommand{\arraystretch}{1.08}
\resizebox{\textwidth}{!}{
\begin{tabular}{lcccccccc}
\toprule
Relation
& DPR
& Spider
& Spider-NQ
& Spider-Trivia
& Contriever
& E5
& Qwen3-4B
& Qwen3-8B \\
\midrule
applies to jurisdiction
& 0.280 / 78.5\%
& 0.361 / 68.1\%
& 0.481 / 85.3\%
& 0.404 / 80.4\%
& 0.198 / 43.3\%
& --
& 0.414 / 100.0\%
& -0.052 / 97.6\% \\

native language
& 0.029 / 0.9\%
& 0.347 / 0.1\%
& 0.451 / 0.5\%
& 0.421 / 1.7\%
& --
& 0.103 / 79.2\%
& 0.095 / 56.8\%
& 0.148 / 88.1\% \\

occupation
& 0.402 / 2.5\%
& 0.399 / 0.0\%
& 0.365 / 0.0\%
& 0.458 / 1.3\%
& --
& --
& -0.087 / 40.2\%
& -0.221 / 72.4\% \\

employer
& 0.148 / 48.9\%
& --
& --
& --
& --
& --
& 0.279 / 71.4\%
& 0.440 / 97.5\% \\

owned by
& 0.445 / 96.4\%
& 0.265 / 20.1\%
& 0.199 / 54.9\%
& 0.323 / 23.3\%
& --
& --
& 0.266 / 56.7\%
& 0.181 / 82.9\% \\

located in administrative territorial entity
& -0.135 / 58.7\%
& 0.249 / 91.9\%
& 0.161 / 58.8\%
& 0.251 / 97.8\%
& --
& --
& 0.161 / 92.3\%
& 0.410 / 98.4\% \\

capital of
& 0.053 / 50.7\%
& 0.156 / 86.3\%
& 0.106 / 61.8\%
& 0.383 / 50.5\%
& --
& --
& 0.301 / 83.5\%
& 0.238 / 97.4\% \\

headquarters location
& -0.244 / 13.5\%
& 0.359 / 95.2\%
& 0.397 / 52.1\%
& 0.352 / 60.9\%
& --
& --
& 0.235 / 30.5\%
& 0.274 / 88.4\% \\

country
& -0.176 / 19.3\%
& 0.213 / 44.6\%
& 0.275 / 6.9\%
& 0.284 / 68.5\%
& --
& 0.214 / 51.6\%
& 0.457 / 83.6\%
& 0.405 / 97.7\% \\

manufacturer
& 0.186 / 100.0\%
& 0.242 / 81.0\%
& 0.189 / 99.3\%
& 0.338 / 98.5\%
& -0.102 / 73.8\%
& 0.354 / 64.1\%
& 0.302 / 99.0\%
& 0.182 / 99.2\% \\

developer
& 0.245 / 99.3\%
& 0.271 / 81.5\%
& 0.298 / 97.6\%
& 0.399 / 96.0\%
& --
& --
& 0.245 / 70.1\%
& 0.205 / 86.6\% \\

place of birth
& 0.211 / 6.2\%
& 0.305 / 64.4\%
& 0.273 / 16.3\%
& 0.171 / 22.0\%
& --
& 0.269 / 6.2\%
& -0.002 / 3.6\%
& 0.193 / 41.1\% \\

twinned administrative body
& --
& --
& --
& --
& --
& 0.130 / 0.0\%
& --
& -- \\

place of death
& 0.035 / 7.5\%
& --
& 0.329 / 54.8\%
& --
& --
& --
& -0.012 / 0.0\%
& 0.070 / 31.2\% \\

record label
& 0.031 / 1.1\%
& 0.293 / 56.2\%
& -0.190 / 5.0\%
& 0.172 / 3.6\%
& --
& --
& 0.132 / 13.3\%
& 0.270 / 32.3\% \\

location
& 0.015 / 20.0\%
& 0.278 / 52.6\%
& 0.200 / 88.1\%
& 0.247 / 48.2\%
& --
& 0.118 / 100.0\%
& 0.061 / 79.3\%
& 0.193 / 93.6\% \\

continent
& -0.020 / 56.0\%
& 0.049 / 55.3\%
& 0.145 / 21.5\%
& 0.089 / 71.6\%
& --
& --
& --
& -- \\

instance of
& 0.295 / 98.4\%
& -0.083 / 82.1\%
& -0.096 / 99.8\%
& -0.111 / 91.7\%
& --
& --
& 0.090 / 100.0\%
& 0.233 / 100.0\% \\

capital
& -0.114 / 7.9\%
& 0.062 / 39.2\%
& 0.148 / 30.7\%
& 0.203 / 20.3\%
& --
& --
& 0.168 / 65.9\%
& 0.115 / 62.3\% \\

part of
& --
& --
& --
& --
& --
& 0.558 / 84.4\%
& --
& -- \\

original broadcaster
& 0.411 / 53.4\%
& 0.126 / 63.6\%
& 0.311 / 58.1\%
& 0.334 / 25.0\%
& --
& --
& 0.197 / 14.0\%
& 0.144 / 48.6\% \\

member of
& 0.375 / 21.3\%
& 0.506 / 85.2\%
& 0.607 / 83.5\%
& 0.585 / 83.1\%
& --
& --
& --
& -- \\

shares border with
& 0.289 / 18.7\%
& 0.598 / 3.0\%
& 0.582 / 37.8\%
& 0.474 / 1.8\%
& --
& -0.194 / 23.1\%
& 0.307 / 0.6\%
& 0.300 / 25.4\% \\

country of origin
& 0.138 / 30.0\%
& 0.289 / 28.9\%
& 0.114 / 61.3\%
& --
& --
& --
& --
& -- \\

has part
& 0.455 / 91.9\%
& 0.401 / 31.3\%
& 0.485 / 81.9\%
& 0.423 / 58.6\%
& --
& 0.339 / 85.3\%
& --
& -- \\

diplomatic relations
& 0.043 / 1.1\%
& 0.308 / 0.0\%
& 0.386 / 1.1\%
& 0.444 / 0.0\%
& --
& --
& -0.004 / 0.0\%
& 0.281 / 0.0\% \\

location of formation
& -0.038 / 8.3\%
& --
& --
& --
& --
& --
& 0.179 / 25.5\%
& 0.246 / 63.9\% \\

work location
& 0.179 / 19.0\%
& 0.493 / 80.0\%
& 0.243 / 92.2\%
& 0.394 / 48.5\%
& --
& 0.021 / 43.3\%
& 0.267 / 10.8\%
& 0.292 / 80.6\% \\
\bottomrule
\end{tabular}
}
\caption{Relation-level \relchange{} results for relation--model pairs
with at least 30 examples, restricted to relations used in
\cref{tab:relation-mappings}. Each cell reports Pearson correlation /
wrong-relation preference rate. Dashes indicate insufficient examples for
that relation--model pair.}
\label{tab:relation-level-relchange}
\end{table*}
\clearpage

\section{Shared Fixed-Example Model-Size Analysis}
\label{app:shared-200}

In the main experiments, examples are selected separately for each model
because the object-knowledge diagnostic \(x_i^R\) is model-specific. This
allows us to test whether larger diagnostic scores predict larger
retrieval-score shifts within each model, but it also means that different
models may be evaluated on different selected examples.

To check whether cross-model trends are driven by model-specific example
selection, we run an additional fixed-example analysis. We construct a shared
set of 200 LAMA/TREx examples using E5-large as the reference model, selecting
20 examples from each of its ten object-knowledge bins in
\(0 \leq x_i^R < 10\). The same source example IDs are then evaluated for all
retrievers and in both the \objchange{} and \relchange{} settings.

\begin{figure*}[t]
\centering
\includegraphics[width=0.78\textwidth]{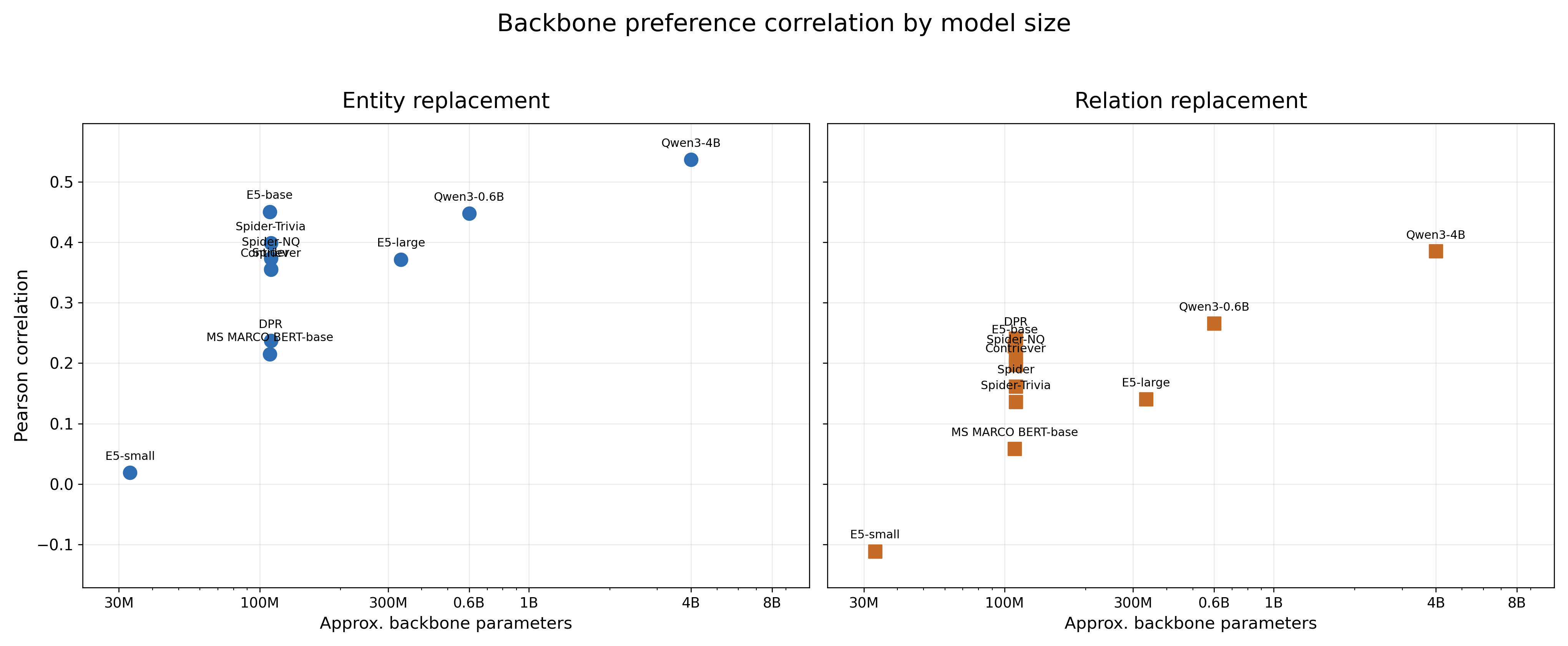}
\caption{Shared fixed-example model-size analysis. All models are evaluated on
the same 200 examples, selected using E5-large as the reference model with 20
examples from each of its ten object-knowledge bins. The plot shows approximate
backbone parameter count against the Pearson correlation between the
model-specific object-knowledge diagnostic and the retrieval-score shift in the
\objchange{} and \relchange{} settings.}
\label{fig:shared-200-model-size}
\end{figure*}

For each model and setting, we recompute the model-specific object-knowledge
diagnostic and retrieval-score differences on the shared set, using z-score
normalization within the shared run. We then compute the Pearson correlation
between the object-knowledge diagnostic and the retrieval-score shift, together
with the corresponding preference rate. Figure~\ref{fig:shared-200-model-size}
shows the relationship between approximate backbone parameter count and Pearson
correlation in both settings.

\section{Original-Backbone Representation Baseline}
\label{app:original-backbone}

Our main diagnostic is computed from the representation produced by the
retrieval-trained model, projected through the original backbone's
token-prediction head. To test whether the observed association is already
present in the original pretrained backbone, we repeat the diagnostic using
the representation produced by the backbone before retrieval training.

Specifically, the main diagnostic is
\[
x_i^{\mathrm{retriever},R}
=
\log p_{M_R}(o_i^{+}\mid u_i)
-
\log p_{M_R}(o_i^{-}\mid u_i),
\]
where \(M_R\) is the retrieval-trained model. For the baseline, we compute
\[
x_i^{\mathrm{backbone},R}
=
\log p_{B_R}(o_i^{+}\mid u_i)
-
\log p_{B_R}(o_i^{-}\mid u_i),
\]
where \(B_R\) is the corresponding original pretrained backbone, before
retrieval training.

Only the representation used to compute the object-knowledge diagnostic is
changed. In both cases, the retrieval-score differences
\(y_i^{\mathrm{obj},R}\) and \(y_i^{\mathrm{rel},R}\) are produced by the
final retrieval-trained model.

We use the same 1,000 examples for both diagnostics. These examples are the
model-specific samples used in the main experiment, selected by requiring
\(x_i^{\mathrm{retriever},R}>0\) and then sampling uniformly at random.
We do not re-select examples according to
\(x_i^{\mathrm{backbone},R}\).
Table~\ref{tab:original-backbone} reports the Pearson correlations obtained
with the retrieval-trained and original-backbone representations.

\begin{table*}[t]
\centering
\small
\begin{tabular}{lrrrr}
\toprule
& \multicolumn{2}{c}{\objchange{}}
& \multicolumn{2}{c}{\relchange{}} \\
\cmidrule(lr){2-3}
\cmidrule(lr){4-5}
Model
& Retriever repr.
& Original backbone repr.
& Retriever repr.
& Original backbone repr. \\
\midrule
DPR           & 0.21 & 0.22 & 0.19 & 0.19 \\
Spider        & 0.29 & 0.24 & 0.22 & 0.24 \\
Spider-NQ     & 0.41 & 0.38 & 0.30 & 0.31 \\
Spider-Trivia & 0.48 & 0.46 & 0.26 & 0.30 \\
Contriever    & 0.29 & 0.13 & 0.13 & 0.20 \\
E5-large      & 0.45 & 0.60 & 0.30 & 0.25 \\
Qwen3-4B      & 0.52 & 0.46 & 0.28 & 0.25 \\
Qwen3-8B      & 0.47 & 0.46 & 0.29 & 0.28 \\
\bottomrule
\end{tabular}
\caption{Pearson correlations between the object-knowledge diagnostic and the
final retriever's score difference. ``Retriever repr.'' computes the
diagnostic from the representation of the retrieval-trained model, whereas
``Original backbone repr.'' computes it from the representation of the
corresponding pretrained backbone before retrieval training. Both correlations
use the same 1,000 examples and the same final-retriever score differences.}
\label{tab:original-backbone}
\end{table*}

As shown in Table~\ref{tab:original-backbone}, correlations remain positive
for every model and setting when the diagnostic is computed from the original
pretrained backbone representation.

\section{QAMPARI Known-vs-Unknown Retrieval Analysis}
\label{app:qampari}

The controlled experiments isolate \biasname{} using template-generated
passage contrasts. We additionally evaluate whether a related pattern appears
in a natural multi-answer retrieval setting using QAMPARI. QAMPARI questions have multiple correct answers, making them useful for testing whether a
retriever favors evidence for an answer that the model already appears to know
over evidence for other valid answers to the same question.

For each model, we select questions with one correct answer whose answer-token
rank is at most 5, which we call the \emph{known} answer, and correct
alternatives from the same question whose answer-token ranks are greater than
20, which we call \emph{unknown} answers. Each valid unknown answer is counted
separately.

We then retrieve passages for the original question and compare where
evidence for the known and unknown answers appears in the top-100 retrieval
results.

The main unit of analysis is a known-vs-unknown comparison within the same
question and retriever. This controls for the question and asks whether
retrieval favors the answer that the model's object-knowledge diagnostic
identifies as more familiar. Pairwise and mean-rank analyses show a known-answer preference for every evaluated retriever. 

\begin{figure*}[t]
\centering
\includegraphics[width=0.82\textwidth]{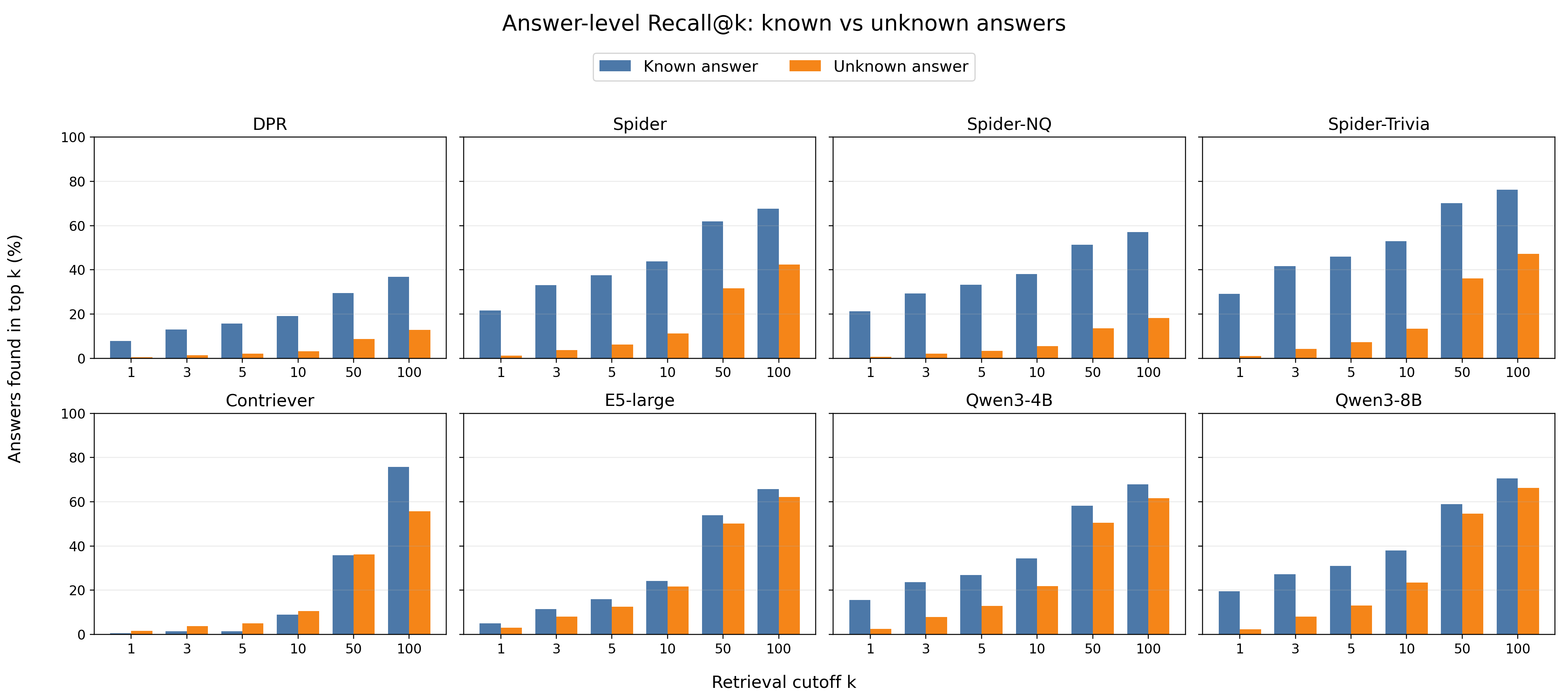}
\caption{Answer-level recall for known and unknown QAMPARI answers. Unknown
answers are counted individually, rather than using an any-unknown-per-question
metric.}
\label{fig:qampari-recall}
\end{figure*}

Figure~\ref{fig:qampari-recall} compares how often known and unknown answers
appear in the top \(k\) results. Each unknown answer is counted separately,
rather than counting whether any unknown answer appears for the question. This
gives a stricter answer-level comparison between the known and unknown sides.

\begin{figure*}[t]
\centering
\includegraphics[width=0.82\textwidth]{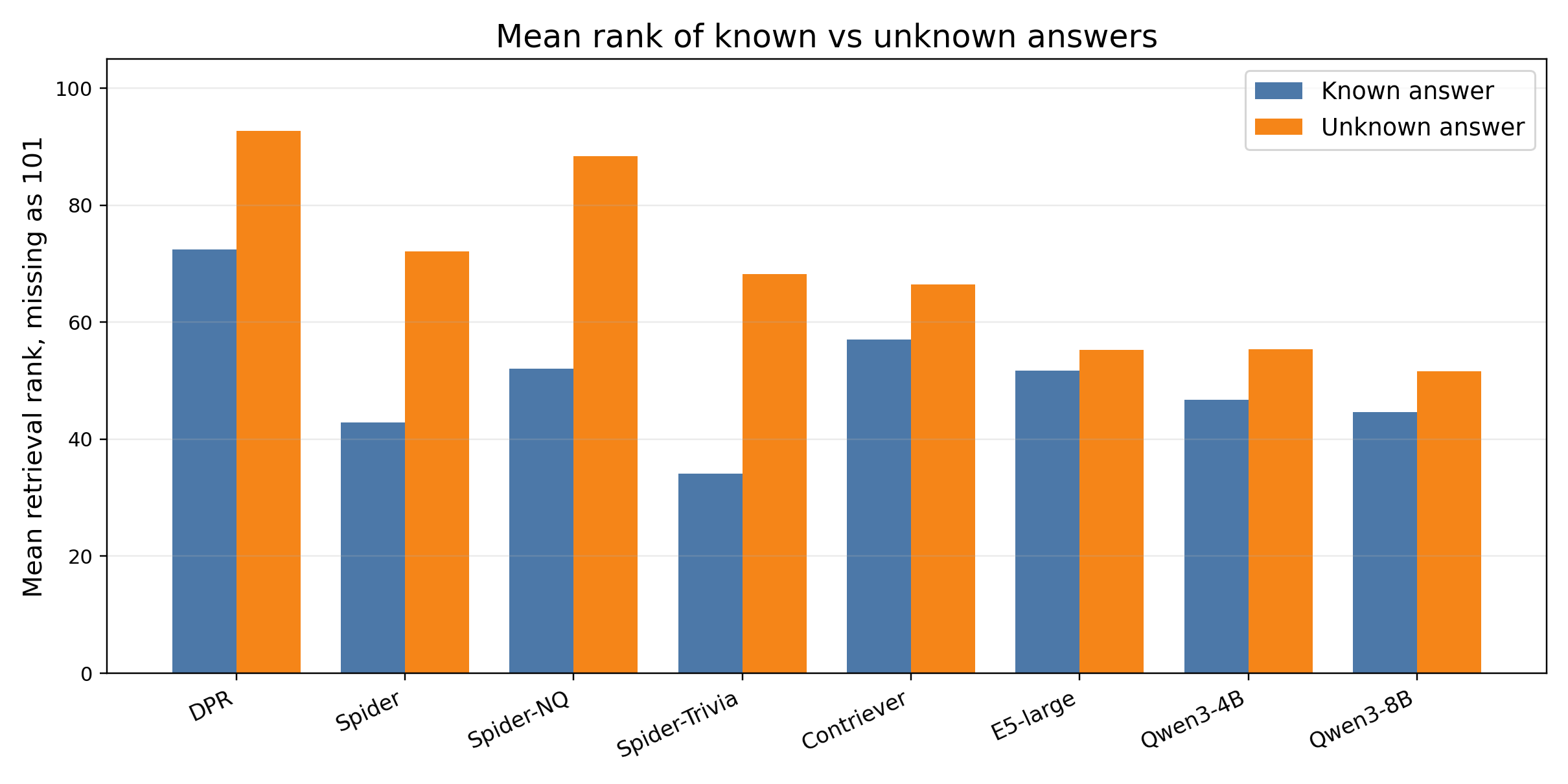}
\caption{Capped rank@100 for known and unknown QAMPARI answers. Answers not
retrieved in the top-100 are counted as one position below the cutoff, rather
than as their true corpus-level rank.}
\label{fig:qampari-mean-rank}
\end{figure*}
Figure~\ref{fig:qampari-mean-rank} reports a capped rank@100 diagnostic for
known and unknown answers. Since the retrieval evaluation is restricted to the
top-100 results, answers not retrieved within this range are counted as one
position below the cutoff. This value should not be interpreted as the true
corpus-level rank. Instead, it provides a bounded penalty for retrieval
failures while keeping the metric focused on the evaluated top-100 range. This
avoids letting a small number of far-tail ranks dominate the mean, while still
distinguishing answers that are retrieved early from answers that are not
retrieved within the cutoff.

This analysis differs from the controlled LAMA/TREx experiments in two main
ways. First, the passages are retrieved from a natural retrieval corpus rather
than generated from templates. Second, QAMPARI contains multiple valid answers
per question, so the question is not whether the canonical answer is correct,
but whether retrieval disproportionately favors the answer that the model
already appears to know.

\section{Passage Annotation Protocol}
\label{app:annotation}

To better understand why top-ranked passages are retrieved in the QAMPARI
analysis, we annotate retrieved passages with answer- and relation-level
labels. The annotation asks whether each passage actually answers the question,
whether it contains the selected known answer, whether it contains one of the
unknown answers, and whether it expresses the same relation requested by the
question.

The annotation is used only as a diagnostic analysis of retrieved passages. It
is not used to compute the main QAMPARI ranking metrics, which are computed
directly from retrieval ranks. Gold proof passages from QAMPARI are used when
available. Non-gold passages are annotated with an LLM using a fixed system
prompt, a fixed user-prompt template, and a restricted set of annotation
classes.

For each retrieved passage, the annotation includes the following fields:

\begin{itemize}
    \item \texttt{known\_answer\_present}: whether the passage contains the
    selected known answer.
    \item \texttt{unknown\_answer\_present}: whether the passage contains one
    of the unknown answers.
    \item \texttt{answers\_question}: whether the passage answers the original
    question.
    \item \texttt{same\_relation\_as\_question}: whether the passage expresses
    the same relation requested by the question.
    \item \texttt{primary\_label}: the main diagnostic category assigned to the
    passage.
    \item \texttt{short\_reason}: a short explanation for the assigned label.
\end{itemize}

The annotation labels distinguish between passages that genuinely support a
known or unknown answer and passages that are only topically related. This is
important because a passage may receive a high score for different reasons: it
may contain the known answer, contain another valid answer, express the correct
relation, mention the same subject or entity type, or simply belong to the same
broad topic.

\section{LLM Annotation Task and Prompt}
\label{app:llm-prompt}

The LLM annotation task is to classify each non-gold retrieved passage
according to its relation to the question, the selected known answer, and the
set of unknown answers. The annotator is instructed to use only the provided
question, answers, and retrieved passage, and not to rely on outside knowledge.
The prompt is designed to avoid treating answer-string overlap as sufficient
evidence that a passage answers the question.

\paragraph{Prompt inputs.}
For each retrieved passage, the annotation prompt includes the original
question, the selected known answer, the list of unknown answers, the retrieved
passage, the retrieved title, the passage rank, and the retrieval score.

\paragraph{Annotation instructions.}
The annotator is instructed to use only the provided question, answers, and
retrieved passage, and not to rely on outside knowledge. The annotator is asked
to determine whether the known answer is present, whether any unknown answer is
present, whether the passage answers the question, whether it expresses the
same relation as the question, and which primary diagnostic label best
describes the passage. The prompt explicitly warns that answer-string overlap
alone is not sufficient: a passage should be marked as answering the question
only when it expresses the relation requested by the question.

\paragraph{Output fields.}
The annotation output contains the following fields:
\texttt{known\_answer\_present}, \texttt{unknown\_answer\_present},
\texttt{answers\_question}, \texttt{same\_relation\_as\_question},
\texttt{primary\_label}, and \texttt{short\_reason}.

\paragraph{Annotation classes.}
The allowed primary labels are:

\begin{itemize}
    \item \texttt{known\_answer\_match}: the passage supports the known answer
    for the question.
    \item \texttt{unknown\_answer\_match}: the passage supports one of the
    unknown answers for the question.
    \item \texttt{answers\_question\_other}: the passage answers the question,
    but not with the selected known answer or listed unknown answers.
    \item \texttt{known\_answer\_present\_wrong\_relation}: the passage
    contains the known answer but does not express the relation asked by the
    question.
    \item \texttt{unknown\_answer\_present\_wrong\_relation}: the passage
    contains an unknown answer but does not express the relation asked by the
    question.
    \item \texttt{same\_topic\_or\_domain}: the passage is topically related
    but does not clearly answer the question.
    \item \texttt{no\_clear\_relation}: the passage does not clearly express
    the relation needed to answer the question.
    \item \texttt{insufficient\_information}: the passage is too ambiguous or
    incomplete to assign a more specific label.
\end{itemize}
\end{document}